\documentclass{article}

\usepackage[english]{babel}
\usepackage[utf8]{inputenc}

\usepackage[letterpaper,top=2cm,bottom=2cm,left=3cm,right=3cm,marginparwidth=1.75cm]{geometry}

\usepackage{amsmath}
\usepackage{graphicx}
\usepackage{mwe}
\usepackage{amssymb}
\usepackage[colorlinks=true, allcolors=blue]{hyperref}
\usepackage[dvipsnames]{xcolor}
\usepackage{xr}
\usepackage{colortbl}
\usepackage{booktabs}
\usepackage[ruled,vlined]{algorithm2e}
\usepackage[none]{hyphenat}
\usepackage{caption}

\definecolor{commentcolor}{RGB}{110,154,155}   
\newcommand{\PyComment}[1]{\ttfamily\textcolor{commentcolor}{\# #1}}  
\newcommand{\PyCode}[1]{\ttfamily\textcolor{black}{#1}} 
\title{Learning to See the Elephant in the Room: \\ Self-Supervised Context Reasoning in Humans and AI}
\date{}

\author {
    Xiao Liu\textsuperscript{\rm 1, 2, *},
    Soumick Sarker\textsuperscript{\rm 1, *}
    Ankur Sikarwar\textsuperscript{\rm 1, 2},
    Bryan Atista Kiely\textsuperscript{\rm 1, 2}, \\
    Gabriel Kreiman\textsuperscript{\rm 3},
    Zenglin Shi\textsuperscript{\rm 1, 2},
    Mengmi Zhang\textsuperscript{\rm 1}\\
    \textsuperscript{\rm 1} \small College of Computing and Data Science, Nanyang Technological University, Singapore\\
    \textsuperscript{\rm 2} \small CFAR and I2R, Agency for Science, Technology and Research, Singapore,\\ 
    \textsuperscript{\rm 3} \small Boston Children's Hospital, Harvard Medical School, USA, \\
    \textsuperscript{\rm *} \small Authors contributed equally to this work.\\
    \small Address correspondence to mengmi.zhang@ntu.edu.sg
}

\usepackage{microtype}

\begin{document}
\maketitle

\begin{abstract}

A tiny object on the table might be a fork, but not an elephant. Humans rarely perceive objects in isolation; instead, they interpret scenes through relationships among co-occurring elements. But how do humans learn these contextual associations? We address this question through a series of human psychophysics experiments. First, we designed a set of context rules by replacing familiar household objects with novel ones. These rules capture different types of associations: global context (e.g., a toothbrush typically appears in bathrooms), local context (e.g., a fork often appears near a plate, even across rooms), and crowding effects (e.g., eggs tend to cluster together). Participants were exposed to short training videos showing these novel objects embedded in naturalistic scenes. We then tested their contextual reasoning using a ``lift-the-flap" task, where the central object was hidden, and participants had to infer its identity based on the surrounding context. We also introduced contextual variations in the task by changing the size, resolution, and spatial arrangement of the scene context. 
Results show that humans can acquire contextual rules in a self-supervised manner without labels or feedback, and can robustly infer the hidden object across a range of contextual variations. 
To model this human capability, we introduce SeCo (Self-supervised learning for Context reasoning), a novel computational model for learning contextual associations. SeCo first identifies candidate target regions, then encodes the target and surrounding context using separate vision encoders. Inspired by semantic memory in biological brains, SeCo includes a learnable external memory module that stores latent contextual priors. Given a contextual cue, SeCo infers the identity of a hidden object by retrieving a likely object representation from this memory and regressing it toward the actual target. In contrast to existing SSL methods that focus on object-centric learning from single-object images, SeCo explicitly learns contextual relationships from complex scenes. Our results show that SeCo outperforms state-of-the-art SSL methods on the lift-the-flap task. Network analysis reveals that its external memory stores meaningful contextual knowledge, enabling accurate inference. Moreover, we also extend the context reasoning ability of SeCo, state-of-the-art SSL methods, and humans to object priming tasks, where they are asked to place target objects in context-appropriate locations. SeCo predicts object placements most closely aligned with human behavior.
In learning to see the elephant in the room, both humans and SeCo reveal that scene understanding arises not from objects alone, but from the contextual associations that bind them together.
\\

\end{abstract}

\textbf{Text statistics}

6 figures

8 supplementary figures





\section{Introduction}

A tiny object on a table might reasonably be a fork, but never an elephant. Object recognition relies not just on the visual features of individual objects, but also on learned context regularities: which objects tend to co-occur, their relative sizes, and their typical spatial relationships. For example, a fork often appears with a knife, is smaller than a vase, and is typically placed to the right of the plate.

While the field of visual neuroscience has long been captivated by the computational challenges of object recognition \cite{riesenhuber1999hierarchical,yamins2016eight,kar2024quest}, most studies have centered on isolated objects \cite{liu2009timing,majaj2015simple,cortex2005fast}.
There has been enduring interest in the statistics of natural scenes and foundational behavioral studies demonstrating the role of context in object recognition \cite{bar2006top,bar2004visual,bar2003cortical,vo2021meaning,palmer1975effects,bar1996spatial,auckland2007nontarget, davenport2004scene,whenpigsfly,puttingincontext}, detection \cite{BIEDERMAN1982143,hollingworth1998does}, and search \cite{chun1998contextual, henderson1999effects}. Yet, how the brain learns to represent high-level contextual priors and integrates them with sensory input to support downstream tasks such as object recognition and detection remains poorly understood.

Prior neuroscience research on contextual modulation has primarily addressed low-level visual processes \cite{fisher2017retinal,henry2020spatial,mely2018complementary,keller2020disinhibitory}, such as extra-classical receptive fields and surround suppression. While foundational, these mechanisms do not account for how high-level contextual knowledge is acquired, represented, or applied. Recent neurophysiological studies \cite{djambazovska2024impact} in macaques have started to explore how high-level context shapes object recognition, yet the developmental origins of this knowledge remain largely unknown. These findings also highlight persistent gaps between human and AI performance in using context; however, there are few concrete strategies proposed for bridging this divide through improved computational models.

In recent years, interdisciplinary research in vision science and AI has made progress in modeling cortical processing pathways using artificial neural networks (ANNs) \cite{hong2016explicit,kar2021fast,majaj2015simple,connor2007transformation, rajalingham2018large, kar2019evidence, rajalingham2015comparison, oshea2015introductionconvolutionalneuralnetworks, resnet, girshick2014rich}. Some of these ANNs partially explain responses in multiple areas in the visual cortex \cite{cadena2019deep,laskar2020deep,yamins2014performance,pospisil2018artiphysiology,bashivan2019neural}, yet they diverge from real-world object recognition in several ways \cite{puttingincontext,geirhos2018imagenet,geirhos2018generalisation,paulun2022distributed}. Humans actively move their eyes to gather context in a task-dependent manner, even when the target’s appearance is uncertain and the surrounding scene varies in structure and resolution \cite{puttingincontext,henderson1999effects}. To incorporate contextual understanding into computational modeling, some computer vision studies have employed statistical optimization techniques \cite{gonfaus2010harmony, yao2012describing, ladicky2010graph, chen2017deeplab}, graph neural networks \cite{battaglia2016interaction, choi2012unified, deng2016structure, hu2016learning}, and transformer-based approaches \cite{carion2020end, whenpigsfly} across tasks such as object recognition \cite{puttingincontext, whenpigsfly}, object detection \cite{liu2018structure, chen2018iterative}, semantic segmentation \cite{mottaghi2014role}, and visual question answering \cite{teney2017graph}. However, most of these models are trained on millions of labeled images \cite{puttingincontext, whenpigsfly}, which is misaligned with how humans learn from natural environments with minimal supervisory signals. 

Evidence has shown that infants in early months form categorical representations of visually similar natural categories without supervision \cite{quinn1993evidence}.
Self-supervised learning (SSL) algorithms have recently emerged as promising alternatives \cite{contextencoder,simclr,simsiam,dino,barlowtwins,vicreg,xie2021unsupervised}. Although some models exhibit closer alignment with neural responses in cortical processing pathways in biological brains \cite{zhuang2021unsupervised}, they primarily focus on isolated single-object images and pay limited attention to inter-object relationships in naturalistic scenes. Investigating how humans acquire and apply contextual information with little supervision, along with the supporting neural mechanisms in neuroscience, provides valuable constraints that can guide the creation of more human-like and context-aware AI systems.

In our work, to address how humans learn contextual associations, we conducted human psychophysics experiments to quantify context learning and generalization in naturalistic scenes. Participants were first exposed to short training videos showing novel objects embedded in scenes following predefined contextual rules. These rules are based on global (e.g., room-level), local (e.g., co-occurrence), and crowding-based associations. We then tested their reasoning in a ``lift-the-flap" task, where the central object was hidden and participants inferred its identity based on context. Despite no labels or feedback, participants rapidly learned these rules and generalized across variations in context area, resolution, and spatial layout.

To model this human capability, we introduce SeCo (Self-supervised learning for Context reasoning), a novel computational model for learning contextual associations. SeCo first identifies candidate target regions, then encodes the target and its surrounding context using separate visual encoders, mirroring the two visual pathways in the brain between ventral regions \cite{ungerleider1982two} (for object identity) and dorsal regions \cite{grill2004human,epstein1998cortical,auger2012retrosplenial,epstein2008parahippocampal} 
(for contextual and spatial layout processing). Inspired by semantic memory systems in the brain \cite{bar2004visual,mcclelland1995there,marr1971simple}, SeCo incorporates a learnable external memory module analogous to the hippocampus and medial temporal lobe \cite{squire2004medial,eichenbaum2000cortical}. This module stores latent contextual priors and retrieves them based on visual cues. In the “lift-the-flap” task, SeCo infers the hidden object from a contextual cue by retrieving its associated memory representation and regressing it toward the target object, loosely echoing hippocampal pattern completion \cite{rolls2013mechanisms}.

SeCo outperforms existing self-supervised baselines on the lift-the-flap task and shows stronger alignment with human behavior. Analysis reveals that its external memory encodes interpretable groupings of contextually related objects. We further evaluate SeCo and humans on an object priming task, where the goal is to place objects in context-appropriate locations. Results show that SeCo predicts placements that closely match human choices. Together, our work demonstrates that contextual scene understanding in both humans and machines emerges from learning the statistical structure that binds objects within its complex environments.

\section{Results} \label{sec:results}

\begin{figure}[!ht]
\begin{center}
\includegraphics[width=14cm]{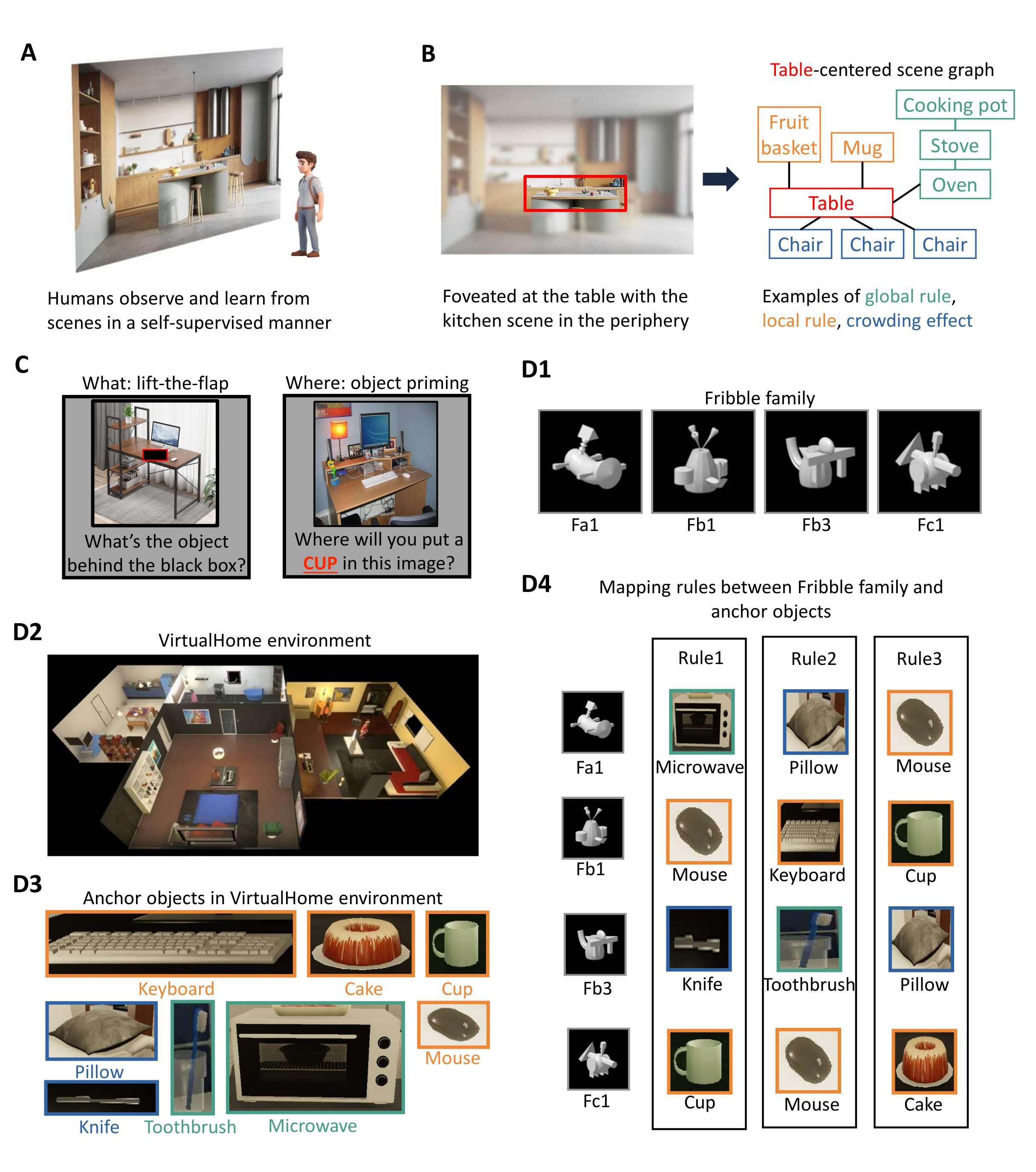}\vspace{-5mm}
\end{center}
\caption{
\footnotesize
\textbf{Humans learn contextual rules from complex natural scenes without explicit instructions or feedback.}  
\textbf{A.}  
Humans perceive scenes holistically rather than as isolated objects. Through exposure to rich, multi-object environments, they implicitly learn contextual associations without the need for explicit instruction or supervision.  
\textbf{B.}  
Foveated vision, with high resolution at the center of gaze and lower resolution in the periphery, supports object-centric representations of complex scenes. When fixating on a table (red), surrounding objects are organized into a table-centered scene graph \cite{khandelwal2023adaptive}. Bounding box colors denote contextual associations: local co-occurrence (e.g., mug on table), global context (e.g., stove in kitchen), and crowding (e.g., multiple chairs near the table).
\textbf{C.}  
To systematically study contextual reasoning in humans and AI, we introduce two evaluation tasks: lift-the-flap and object priming. In lift-the-flap (left), agents use scene context to infer the identity of a hidden object behind a black patch framed in red. In object priming (right), given a scene and a target object that is not already present, agents predict contextually appropriate locations for placing the object.  
\textbf{D.} 
FRibble In the sceNE (FRINE) dataset for studying how humans learn to reason from context.  
Without relying on prior contextual knowledge of familiar objects, we construct the FRINE dataset using novel objects. We begin by selecting four novel object families (Fa1, Fb1, Fb3, Fc1) from the Novel Object Dataset (NOD) \cite{singh2023learning}. These “fribbles” have distinct body structures and appendages that are unfamiliar to humans as shown in \textbf{D1}.  
Next, we replace eight common household object classes, which serve as anchor objects in the Unity-based indoor scene simulator VirtualHome \cite{puig2018virtualhome}, with fribbles. \textbf{D2} shows an example VirtualHome apartment, alongside example images of the eight household objects in \textbf{D3}, each color-coded according to the global, local, and crowding associations defined in \textbf{B}.  
The mapping between anchor objects and fribbles defines the three novel contextual rules, as illustrated in \textbf{D4}. Each column represents a contextual rule, and each row shows the fribble assigned to the corresponding anchor object class within that rule. See \textbf{Methods} for details on FRINE dataset.
}
\vspace{-6mm}
\label{fig:fig1intro}
\end{figure}

We designed a two-stage experiment for humans and AI models: a training stage (learning to reason or  LoR) and a test stage to measure their abilities to reason from varying contexts (\textbf{fig.~\ref{fig:fig1intro}D4}). 
For the test stage, we introduced two tasks, lift-the-flap, and object priming (\textbf{Fig.~\ref{fig:fig1intro}C}), to address ``what'' and ``where'' questions in context reasoning. During the LoR training stage, human subjects and AI models are tasked with learning contextual associations from the visual input of scenes, enabling them to understand relationships between objects and their environments for effective reasoning in subsequent tasks. In the test stage, the lift-the-flap task evaluates how humans and AI models use scene context to infer the category of a hidden object obscured by a black patch, relying solely on surrounding contextual cues. In a complementary manner, the object priming task assesses the ability to predict contextually appropriate regions within a scene to place a target object that is not initially present. These tasks provide a framework for testing contextual reasoning during inference across different scenarios. See \textbf{Methods} for details.

To minimize biases stemming from human familiarity with real-world scenes, we curated a scene dataset, \textit{FRIbble in the sceNE} (FRINE). FRINE features novel contextual associations by replacing novel objects named \textit{fribble} \cite{singh2023learning} (\textbf{Fig.~\ref{fig:fig1intro}D1}) with existing objects in the VirtualHome environment \cite{puig2018virtualhome}(\textbf{Fig.~\ref{fig:fig1intro}D2}).
We defined a set of replacements between a \textit{fribble} and a VirtualHome object as a \textit{context rule}. We designed three distinct \textit{context rules} to evaluate the reasoning abilities of humans (\textbf{Fig.~\ref{fig:fig1intro}D4}, see \textbf{Methods} for detailed context rules). To assess the robustness of human and AI models' performance on lift-the-flap, we introduced three context manipulations on the FRINE dataset (blur, reduced, and jigsaw context) (\textbf{Fig.~\ref{fig:fig1intro-human}C}). 
We designed a two-stage human psychophysics experiment: a training stage (LoR) and a test stage (lift-the-flap, object priming) to measure human learning and reasoning from varying contexts (\textbf{Fig.~\ref{fig:fig1intro-human}A}, see \textbf{Methods} for details). 
Specifically, subjects were initially exposed to 40 training video clips, each comprising 15 to 20 frames that rotate around the replaced \textit{fribble} and are presented only once. Each frame features the \textit{fribble} alongside its surrounding objects. After the training phase, subjects view 40 test video clips where the replaced \textit{fribble} is hidden behind a black box mask. Each test video clip is shown for an unbounded duration, allowing subjects to observe the context and make a decision regarding the identity of the hidden fribble. 
To investigate whether humans can learn to reason from context in a self-supervised manner, we introduced two training modes in LoR (\textbf{Fig.~\ref{fig:fig1intro-human}ABD}). We refer to human participants trained via supervised learning as ``SUP humans" and participants trained via self-supervised learning as ``SSL humans" for short throughout the rest of the text. In lift-the-flap, we tested 160 human subjects in online experiments recruited through Amazon Mechanical Turk (AMT), randomly assigning participants to different training modes, context rules, and context manipulations, with appropriate quality control measures in place. 
In the object priming task, we recruited 437 human subjects in online experiments through AMT. See \textbf{Methods.} for details.

To investigate whether capturing contextual associations and consequently leveraging them during reasoning is a common capability of self-supervised learning (SSL) methods in AI, we considered several state-of-the-art baselines: Context Encoder \cite{contextencoder}, SimCLR \cite{simclr}, SimSiam \cite{simsiam}, DINO \cite{dino}, VICReg \cite{vicreg}, ORL \cite{xie2021unsupervised}. We further introduced a self-supervised learning method for context reasoning, named SeCo, which learns associations between objects and their surrounding contexts in scene images (\textbf{Fig.~\ref{fig:fig2arch}A}). When humans fixate on an object in a complex scene, they simultaneously perceive its surrounding objects in their peripheral vision (\textbf{Fig.~\ref{fig:fig1intro}A}). Similarly, SeCo incorporates an object-context discovery module, using an unsupervised region proposal method (selective search \cite{selectivesearch}) to locate potential objects of interest and features a two-stream visual processor, where objects are processed in high resolution while their surrounding contexts are processed at a lower resolution. As humans learn and memorize contextual associations after exposure to scenes, SeCo employs an external memory during the pre-training to simulate this underlying mechanism of context modulation. We evaluated SeCo and SSL baselines on both lift-the-flap and object priming tasks after fine-tuning. We curated two sets of natural scene datasets, COCO-OCD and COCO-VOC, to study AI models' reasoning ability in more complex contexts and their robustness against domain shifts. To ensure fair comparisons, all methods are initialized with ImageNet-pretrained weights \cite{imagenet}, followed by pre-training and finetuning on the in-domain training set (\textbf{Fig.~\ref{fig:fig2arch}B}), then evaluated on the in-domain test set and in a zero-shot setting on the out-of-domain datasets (\textbf{Fig.~\ref{fig:fig2arch}C}). See \textbf{Methods} for detailed curation process and evaluation protocols. 



\subsection{Human and AI can learn to reason from context without supervisions
}\label{sec:results-human-general}





\begin{figure}[!ht]
\begin{center}
\includegraphics[width=14cm]{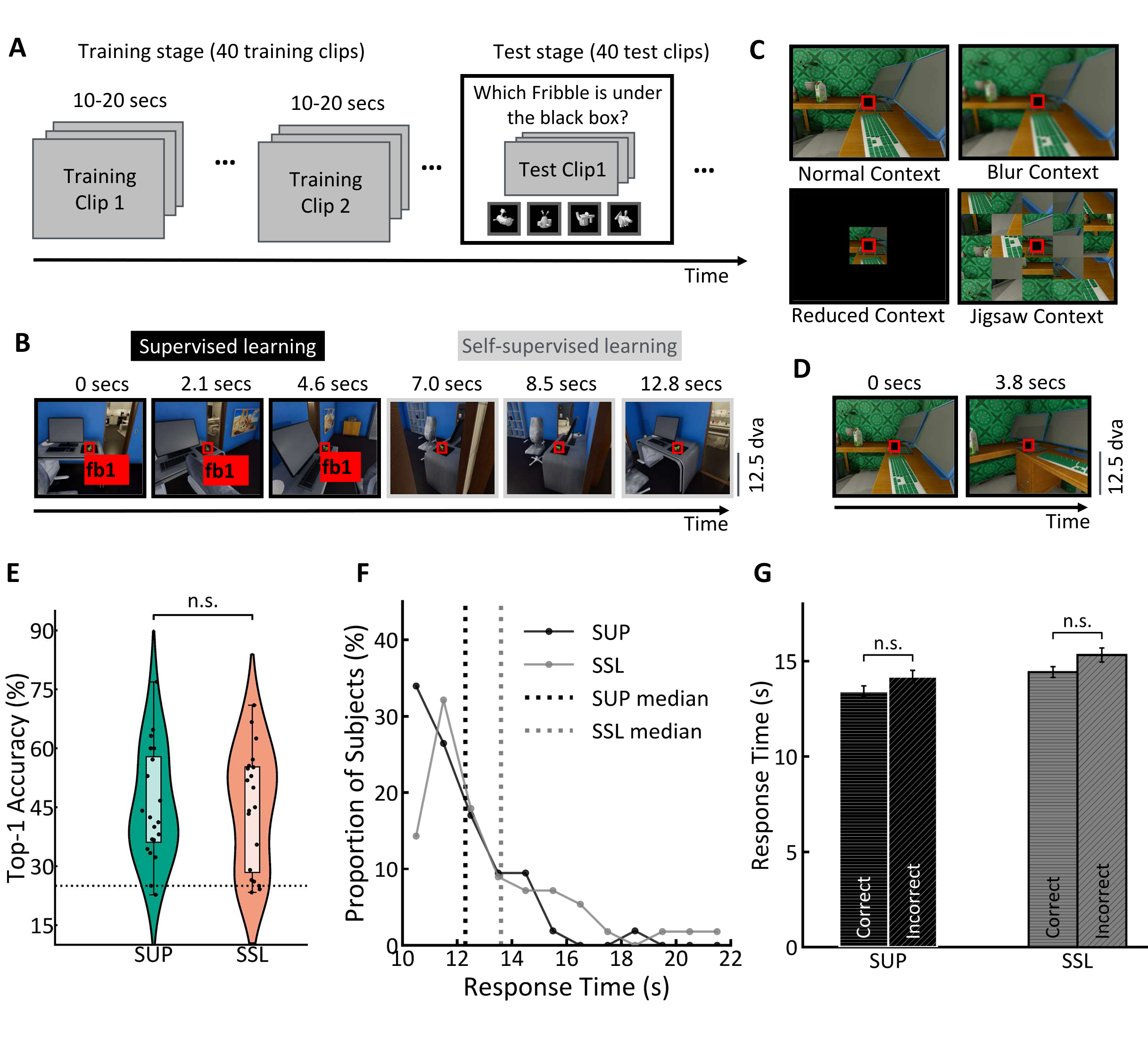}\vspace{-9mm}
\end{center}
\caption{
\footnotesize
\textbf{We introduce human psychophysics experiments in learning to reason on the FRINE dataset.} 
\textbf{A.} Schematic of the Human Psychophysics Experiments. The experiment comprises two phases: a training phase and a testing phase. In the training phase, participants were shown a sequence of 40 training video clips from the FRINE dataset. Each clip lasted 10 to 20 seconds and depicted a novel fribble object centered within a naturalistic scene. The camera rotated around the fribble object in each video. See \textbf{B} for an example training clip. Participants were assigned to either a supervised (Sup) or self-supervised learning (SSL) condition. In the Sup condition (clips framed in black), the fribble object was highlighted with a red bounding box and labeled (e.g., “fb1”). In the SSL condition (clips framed in gray), no labels were shown. Importantly, each participant viewed training videos from only one learning condition (either Sup or SSL) throughout the entire training phase. In \textbf{B}, the first three frames are shown in the Sup condition and the remaining frames in SSL for illustration only. Frame timestamps are shown above each frame, and a scale bar on the right indicates the frame size in degrees of visual angle. During the test phase, participants viewed 40 test clips from the FRINE dataset. In each clip, the fribble object was occluded by a black patch, and participants were required to infer its identity based solely on contextual information, making a 4-way forced-choice classification. \textbf{C.} Context Manipulations in the Lift-the-Flap Task. To investigate the role of context, we introduced four context variations in the test clips: normal context, blur context, context areas, and jigsaw context. \textbf{D.} Example Test Clip. We show an example test clip under the normal context condition, where the central fribble object is hidden beneath a black patch while the surrounding context remains intact. Frame timestamps and video frame sizes (in degrees of visual angle) are annotated similarly to \textbf{B}. \textbf{E.} Individual subject performance differences in the lift-the-flap task. Violin plots show the distribution of top-1 accuracy across individual participants under SUP (green) and SSL (orange) learning conditions, with each black dot representing a single subject. The box spans the interquartile range (25th–75th percentiles); whiskers denote the full range of the data.
\textbf{F.} Response time across human subjects for SUP and SSL conditions. The plot shows the proportion of SUP (black) and SSL (dark grey) participants as a function of their mean reaction time per trial (in seconds). Vertical dotted lines indicate the median response time for each group.
\textbf{G.} Reaction times for correct and incorrect trials in SUP and SSL humans. Trials were separated into correct and incorrect responses for SUP and SSL participants, and their average reaction time was computed for each learning condition. Error bars indicate the standard error of the mean (SEM). “n.s.” indicates no statistically significant difference between the two distributions ($p > 0.05$).}
\vspace{-6mm}
\label{fig:fig1intro-human}
\end{figure}

\noindent \textbf{Humans can learn and reason from context.}
We first evaluated human performance on the curated FRINE dataset. Human learning ability varied substantially across individuals (\textbf{Fig.~\ref{fig:fig1intro-human}E}), consistent with pronounced inter-individual variability. Despite these differences, the top-1 accuracy of humans in the self-supervised learning (SSL) condition significantly exceeded that of a chance-level model, indicating that humans can effectively learn contextual associations and leverage them for reasoning even in the absence of explicit supervision (\textbf{Fig.~\ref{fig:fig1intro-human}}; SSL humans vs. chance, Welch's two-tailed t-test, $p<0.05$, $t=229.0$, $df=1{,}998$). Nevertheless, SSL human performance remained well below ceiling, underscoring the intrinsic difficulty of the task and the challenge of reasoning based solely on contextual cues. The mean reaction time per trial for SSL humans was 13.07 $\pm$ 2.52 seconds (\textbf{Fig.~\ref{fig:fig1intro-human}F}). We further examined reaction time distributions separately for correct and incorrect trials (\textbf{Fig.~\ref{fig:fig1intro-human}G}). The mean reaction time was 13.98 $\pm$ 3.60 seconds for correct trials and 14.77 $\pm$ 4.33 seconds for incorrect trials. A significant difference in response times was observed between correct and incorrect trials (Welch's two-tailed t-test, $p<0.05$, t = -2.5, df $\approx$ 621), which agrees with prior object recognition studies in which correct responses are typically faster \cite{he2016deep}.

To establish an upper bound on human performance under supervised learning (SUP humans), we provided explicit labels for fribbles during the LoR training stage (Fig.~\ref{fig:fig1intro}D4) and subsequently tested participants on the lift-the-flap task. As in the SSL condition, learning and reasoning performance varied across individuals; however, average top-1 accuracy for SUP humans was only marginally higher than that of SSL humans (\textbf{Fig.~\ref{fig:fig2arch}D}; SSL vs. SUP humans, Welch's two-tailed t-test, $p<0.05$, $t=3.9$, $df=1{,}998$). This modest performance gain suggests that, for humans, self-supervised learning can be nearly as effective as supervised learning for acquiring and exploiting contextual information for reasoning. The mean reaction time per trial for SUP humans was 13.84 $\pm$ 3.70 seconds (\textbf{Fig.~\ref{fig:fig1intro-human}F}), which was significantly shorter than that observed for SSL humans (Welch's two-tailed t-test, p = 0.002, t = -3.1, df = 621). Reaction time distributions for correct and incorrect trials are shown in \textbf{Fig.~\ref{fig:fig1intro-human}G}. As in the SSL condition, correct trials were associated with slightly shorter mean reaction times (14.39 $\pm$ 3.78) than incorrect trials (15.30 $\pm$ 4.56); however, this difference did not reach statistical significance (Welch's two-tailed t-test, p = 0.057, t = -2.0, df $\approx$ 332).





To ensure data quality and participant attentiveness, we included a memorization test after each video presentation during the training phase, in which subjects were asked whether they had seen the exact same object instance in the preceding video (\textbf{Methods}). As shown in \textbf{Fig.~\ref{fig:figs-memorization}}, 145 out of 160 human subjects using VirtualHome, and 156 out of 160 using ImageNet passed all data quality control criteria based on the memorization tests. To assess whether attentiveness influenced learning performance in SUP humans and SSL humans, we examined the relationship between memorization test performance and top-1 accuracy in the lift-the-flap task. We did not observe a strong correlation between the two (Spearman $\rho=0.12$), suggesting that while most participants were attentive during training, learning ability in this task extends beyond mere visual attentiveness.

Second, we compared online participant performance with that of an in-laboratory cohort (\textbf{Methods}). To further enhance human performance, we increased training duration and introduced a structured curriculum in the in-lab setting. We report top-1 accuracy separately for online and onsite participants under both SUP and SSL conditions (\textbf{Fig.~\ref{fig:figsonsite}}). Onsite participants consistently outperformed the chance model and exceeded online participants by a substantial margin of 9.7\% in top-1 accuracy. These results further confirm that humans can learn to reason from contextual information under both SUP and SSL regimes. The superior performance observed in the laboratory setting is likely attributable to extended training and improved curriculum design.\\


\noindent \textbf{AI models can learn and reason from context with SeCo being the best.}
Similarly, we evaluated SSL methods on the lift-the-flap task using the FRINE dataset. Notably, none of the SSL methods had prior exposure to the FRINE dataset. Despite this, it is remarkable that these methods can transfer contextual knowledge learned from familiar objects to infer the identities of novel objects that share the same contextual rules, outperforming the chance model in a zero-shot setting (\textbf{Fig.~\ref{fig:fig2arch}D}, Welch's two-tailed t-test, $p<0.05$). This result indicates that SSL methods can generalize learned contextual associations to novel objects governed by the same contextual rules without task-specific fine-tuning. Several SSL models (SeCo, ORL, and SimSiam) outperformed SSL humans (Welch's two-tailed t-test, $p<0.05$) as well as the supervised baseline (Welch's two-tailed t-test, $p<0.05$), highlighting their effectiveness in capturing and leveraging contextual structure without explicit labels. However, similar to SSL humans, the performance of SSL methods remains far from perfect, indicating substantial room for improvement despite their strong generalization capabilities.



Importantly, SeCo outperformed both SUP humans and SSL humans, as well as all baseline methods, demonstrating its superior ability to reason from context (\textbf{Fig.~\ref{fig:fig2arch}D}; for all pairwise comparisons between SeCo and humans or other methods, Welch's two-tailed t-test, $p<10^{-3}$). The Context Encoder \cite{contextencoder}, trained with pixel-level reconstruction, showed inferior performance compared to SeCo and other baselines, suggesting that pixel-level reconstruction emphasizes visual details while neglecting local contextual associations (for all pairwise comparisons between the Context Encoder and other models, Welch's two-tailed t-test, $p<0.05$). Contrastive methods such as SimCLR \cite{simclr} were outperformed by non-contrastive approaches like SimSiam \cite{simsiam}, VICReg \cite{vicreg}, and DINO \cite{dino}, likely due to the challenge of selecting effective negative samples in contexts with multiple co-occurring objects (for all pairwise comparisons between SimCLR and other models, Welch's two-tailed t-test, $p<0.05$). 

Moreover, SeCo surpassed ORL \cite{xie2021unsupervised}, a patch-wise SSL method that relies on augmented views of the same object instances from the same input image or different object instances from similar contextual images. This suggests that SeCo is more effective in context reasoning by leveraging an external memory system to retrieve object representations associated with relevant contextual cues. This mechanism guides the context encoder in SeCo to learn accurate context associations, enabling reliable retrieval of target representations from memory. By introducing external memory, SeCo transitions from object-centric to object-context associative representation learning, thereby enhancing context reasoning. Notably, SeCo also outperforms the supervised learning baseline, demonstrating that object-context associative representations can be learned without supervision.\\

\noindent \textbf{Human and SeCo performance vary across context rules and context associations.}
Different context rules impose distinct relational and inferential demands. Examining human and model reasoning across these rules allows us to probe how context reasoning strategies adapt under varying levels of structural complexity.
Consistent with this, both humans and AI models showed significant variation in context reasoning performance across rules (\textbf{Fig.~\ref{fig:fig2arch}E}, one-way ANOVA, $p<0.05$). In Rule 1, humans and AI models achieved comparable performance (\textbf{Fig.~\ref{fig:figs-glc}A}). However, in Rule 2, SUP humans outperformed SSL humans ($p<0.05$, Welch's two-tailed t-test, $t=30.6$, $df \approx 1,998$), whereas the reverse trend was observed in Rule 3 ($p<0.05$, Welch's two-tailed t-test, $t=-22.5$, $df \approx 1,998$).
Notably, SeCo consistently outperformed both groups of human subjects and other AI models in Rules 2 and 3 (all comparisons, Welch's two-tailed t-test, $p<0.05$), demonstrating its robustness in handling the varying complexity of these context rules. Despite these variations, both humans and SeCo maintained above-chance top-1 accuracy across all rules.



Different objects are associated with varying contextual cues, rendering some objects easier to reason about than others. To examine how context association types influence reasoning ability, we categorized virtual home objects into three groups—global, local, and crowding—based on the nature of their contextual associations (\textbf{Methods}). For example, microwaves (global) typically appear in kitchens, providing a strong and unambiguous contextual cue, whereas cups (local) can occur across diverse environments, making context-based reasoning more challenging (\textbf{Fig.~\ref{fig:fig1intro}B}).

Across association types, SUP humans, SSL humans, and SeCo all exhibited higher top-1 accuracy for targets with strong global associations compared to those with local associations (for all comparisons, Welch's two-tailed t-test, $p<0.05$). A similar pattern was observed across all model baselines (ORL, VICReg, SimSiam, Context Encoder, and Supervised), with each model performing significantly better on global than on local associations (\textbf{Fig.~\ref{fig:figs-glc}}; for all within-model comparisons, Welch's two-tailed t-test, $p<0.05$).

Interestingly, both SUP and SSL humans achieved higher performance under the crowding condition than under global or local association types, suggesting a human advantage in leveraging dense contextual structure. This trend did not generalize to all AI models; nevertheless, SeCo consistently outperformed other AI models in the crowding condition. Notably, despite these variations across association types, both humans and SeCo maintained above-chance top-1 accuracy across all context association categories.

\subsection{Contextual variations influence reasoning ability in humans and AI} \label{sec:results-contextconditions}
To evaluate how well humans and AI models can maintain their reasoning abilities when typical visual and spatial cues are disrupted, we analyzed their performance of the lift-the-flap task on the FRINE dataset with the normal context and the manipulated contexts, including blurred, reduced, or jigsaw context (\textbf{Fig.~\ref{fig:fig1intro-human}C}).\\




\noindent \textbf{Blurred context is sufficient for reasoning.}
Blurring the contextual region removes fine-grained visual details while preserving the global scene structure, allowing us to test whether humans and models rely primarily on low-level visual features or higher-level semantic cues for context reasoning. To quantify the effect of context resolution, we applied zero-mean Gaussian blur to the context with standard deviations of $\sigma = 2, 4, 8, 16,$ and $32$ pixels (video resolution: $1024 \times 1280$ pixels). Each subject experienced all blurring conditions, but never for the same target objects within the same scene, preventing learning, memory, and adaptation effects from confounding performance across conditions.

Human top-1 accuracy decreased as contextual blur increased from $\sigma=0$ (unblurred) to $\sigma=32$ (\textbf{Fig.~\ref{fig:figs-conditions}A}; one-way ANOVA, $p<0.05$). Interestingly, even heavily blurred context remained sufficient for successful reasoning in the lift-the-flap task, with human performance consistently exceeding chance across all blur levels (\textbf{Fig.~\ref{fig:fig4cond}A}; Welch's two-tailed t-tests against chance, $p<0.05$). These results align with prior findings showing that global scene structure can support in-context object recognition despite substantial loss of fine-grained visual detail in the background \cite{puttingincontext}.


Specifically, SSL humans performed on par with SUP humans across almost all blur levels (\textbf{Fig.~\ref{fig:fig4cond}A}; Welch's two-tailed t-test, $p<0.05$). Likewise, SeCo effectively reasoned from blurred contexts across all degrees of blurring (Welch's two-tailed t-tests against chance, t = 800.0, $p<0.05$, df $\approx$ 1,947), although performance showed a dramatic decline under extreme blurring conditions. Most SSL models (SeCo, ORL, SimSiam, VICReg, and DINO) consistently outperformed the supervised model under mild to moderate contextual blur (\textbf{Fig.~\ref{fig:figs-conditions}A}, $\sigma \leq 16$; for all comparisons, Welch's two-tailed t-test, $p<0.05$). Performance degraded substantially under extreme blur ($\sigma=32$), leaving SeCo as the only SSL method that continued to outperform the supervised model (\textbf{Fig.~\ref{fig:figs-conditions}A}; for all comparisons, Welch's two-tailed t-test, $p<0.05$). This robustness can be attributed to SeCo’s two-stream visual processing architecture, which encodes contextual information at low resolution while preserving high-resolution representations of target objects. This design mirrors the human periphery–fovea division in visual processing and supports reliable reasoning under extreme blurring by prioritizing global contextual cues over fine-grained visual details.\\

\noindent \textbf{Contextual areas matter for reasoning.}
In reduced-context scenarios, portions of the scene are removed or substantially minimized, leaving less information for both humans and models to infer object identity. We quantified the amount of available context using the context–object (CO) ratio \cite{puttingincontext}, with values of 128, 16, 8, 4, and 2, to assess how effectively humans and models reason under limited contextual information (\textbf{Fig.~\ref{fig:fig4cond}B}). 

Both humans and SSL models were able to reason from limited context, performing significantly above chance across all CO ratios (\textbf{Fig.~\ref{fig:fig4cond}B}, \textbf{Fig.~\ref{fig:figs-conditions}B}; for all sub-conditions, two-tailed t-test against chance, $p<10^{-3}$). However, progressively reducing contextual information led to a significant decline in performance for both humans and SSL models (for all comparisons, one-way ANOVA, $p<0.05$). Under extreme contextual reduction (CO = 2 and 4), SSL humans significantly outperformed SUP humans (for all comparisons, Welch's two-tailed t-test, $p<0.05$). Similarly, SeCo was more robust than all other AI models, including the supervised baseline, under these minimal context conditions (for all comparisons, Welch's two-tailed t-test, $t=522.0$, $p<0.05$, df $\approx$ 1,954).\\

\noindent \textbf{Context reasoning relies on spatial configuration.}
Jigsaw context disrupts the spatial coherence of a scene by scrambling the locations of surrounding objects while preserving the same set of visual elements. This manipulation challenges the extent to which humans and models rely on spatial relationships and scene layout for context reasoning. Specifically, we partitioned the context into $3\times3$, $5\times5$, and $7\times7$ grids and randomly permuted the pieces, while keeping the central patch containing the target object intact (\textbf{Fig.~\ref{fig:fig4cond}C}).

Context reasoning performance for both humans and AI models was strongly affected by spatial configuration (\textbf{Fig.~\ref{fig:fig4cond}C}, \textbf{Fig.~\ref{fig:figs-conditions}C}; one-way ANOVA, $p<0.05$). However, both SUP and SSL human performance under the $3\times3$ and $5\times5$ jigsaw conditions did not differ significantly from the unscrambled context condition, likely because each larger patch retained sufficient local contextual information and because contextual influence decreases with distance from the target \cite{gupta2021visual} (Welch's two-tailed t-test, $p<0.05$). Overall, humans were more robust to spatial disruption than AI models. Among all the AI models, SeCo outperformed baseline methods under smaller-scale spatial disruptions ($3\times3$ and $5\times5$), but its performance degraded under highly fragmented $7\times7$ contexts (for all comparisons, Welch's two-tailed t-test, $p<0.05$).\\

\noindent \textbf{SeCo behaviors align closely with humans across contextual variations.}
To assess whether SeCo captures not only overall performance but also human-like sensitivity to different forms of contextual manipulation, we examined the alignment between model and human behaviors across multiple context conditions. Various SSL models (SeCo, ORL, and VICReg) exhibited strong linear correlations with SSL human performance across four contextual conditions (all $r \geq 0.8$; \textbf{Fig.~\ref{fig:fig4cond}D}). Notably, SeCo achieved higher absolute accuracy than ORL while maintaining comparably high correlations with human performance. Despite these overall similarities, SeCo and SSL humans showed complementary strengths across context manipulations: SeCo outperformed SSL humans under reduced-context conditions, whereas SSL humans achieved higher accuracy under blur and jigsaw contexts.

\subsection{SeCo generalizes context reasoning in natural scenes}
\label{sec:results-naturaldatasets} 
To further assess the reasoning capability of SeCo and the rest of SSL methods, we considered the real-world scenario where scenes are inherently more complex, containing diverse objects, environments, and spatial relationships.
First, we evaluated the performance of SSL methods against the chance model on the in-domain sets of COCO-OCD and COCO-VOC (\textbf{Fig.~\ref{fig:fig3analysis}A1, A3}). Our results show that all SSL methods successfully reason from complex, real-world scenes (for all pairwise comparisons between SSL methods and the chance model, Welch's two-tailed t-test, $p<0.05$). Among them, SeCo and ORL consistently outperform the supervised baseline, highlighting their effectiveness in capturing meaningful associations in diverse environments without supervision (Welch's two-tailed t-test, $p<0.05$). Consistent with observations from the FRINE dataset, SeCo surpasses ORL by a larger margin on both in-domain COCO-OCD and COCO-VOC datasets (Welch's two-tailed t-test, $p<0.05$). These results suggest that SeCo demonstrates a superior ability to learn and reason from context across diverse and complex naturalistic environments.





Next, we evaluated whether SSL methods remain robust to domain shifts in visual features during context reasoning tasks—for example, recognizing a bird flying in the sky regardless of whether the scene is depicted in Picasso or Monet styles. We evaluated all models on out-of-domain OCD test of COCO-OCD and PASCAL-VOC07 test of COCO-VOC in a zero-shot setting without fine-tuning. Compared to the chance model, all SSL methods retained their reasoning ability under domain shifts, with SeCo consistently outperforming all baselines across both out-of-domain datasets (\textbf{Fig.~\ref{fig:fig3analysis}A2, A4}; Welch's two-tailed t-test for all SeCo and baseline comparisons, $p<0.05$). In particular, performance on OCD test was still comparable to in-domain COCO-VOC for all SSL methods (\textbf{Fig.~\ref{fig:fig3analysis}A4}; one-way ANOVA, $p<0.05$), likely because the in-domain and out-of-domain COCO-VOC datasets consist of natural images. In contrast, the shift from natural images in in-domain COCO-OCD to synthetic images in out-of-domain OCD test led to a substantial performance drop across all models (\textbf{Fig.~\ref{fig:fig3analysis}A2}; one-way ANOVA, $p<0.05$). Importantly, SeCo was the least affected by this domain shift, highlighting its ability to capture contextual associations beyond superficial visual feature correlations.

\subsection{Dissecting SeCo reveals critical components for context-based reasoning} \label{sec:results-seco-analysis}

\noindent \textbf{SeCo learns meaningful contextual associations via selective search.}
Objects play a critical role in context reasoning \cite{ding2022efficient, whenpigsfly}. To facilitate the learning of object–object and object–context associations, we introduce a context–object pair discovery module that leverages selective search \cite{selectivesearch} to propose regions likely to contain objects of interest (\textbf{Methods}).

To assess the effectiveness of this discovery module, we replaced object–context image pairs generated by selective search with pairs derived from annotated ground-truth bounding boxes. Surprisingly, SeCo achieved higher performance with selective search than with ground-truth pairs (\textbf{Fig.~\ref{fig:fig3analysis}B, Ground Truth}), suggesting that selective search proposals preserve richer or more diverse contextual cues than tightly localized ground-truth annotations.
To further test whether selective search yields more informative contextual associations than random pairings, we replaced selective search proposals with randomly generated object–image pairs, differing only in the randomization of target region coordinates. As expected, random generation serves as a lower bound for the discovery module (\textbf{Fig.~\ref{fig:fig3analysis}B, Random Generation}), highlighting the importance of region “objectness” for learning meaningful contextual associations.
We also compared selective search with the widely used random cropping strategy from the SSL literature, in which the input image is duplicated and randomly cropped. Replacing selective search with random cropping resulted in a clear performance drop, indicating that selective search’s object-centric region proposal mechanism is more effective for learning contextual associations (\textbf{Fig.~\ref{fig:fig3analysis}B, Random Cropping}).

Finally, to examine the role of selective search during pre-training, we analyzed both the quantity and quality of its region proposals (\textbf{Methods}). Increasing the Intersection over Union (IoU) threshold with ground-truth bounding boxes did not improve performance, whereas including a larger number of proposals led to modest gains (\textbf{Fig.~\ref{fig:fig3analysis}C}). This suggests that proposal diversity and coverage, rather than precise spatial localization, are more critical for effective context learning, providing a plausible explanation for why selective search outperforms ground-truth bounding boxes in SeCo.\\

\noindent \textbf{SeCo stores context associations in the external memory.}
To study the role of external memory in contextual reasoning, we removed the external memory module from the default SeCo model (see \textbf{Methods}). Eliminating external memory resulted in a 12.5\% drop in top-1 accuracy, highlighting the importance of external memory for learning object–context associations. 

We further varied the number of memory slots (\textbf{Fig.~\ref{fig:memory_slots}A}) from 100 to 800, observing a moderate improvement of 2.5\% in top-1 accuracy on the lift-the-flap task. Next, we examined the effect of memory slot dimensionality on lift-the-flap performance (\textbf{Fig.~\ref{fig:memory_slots}A}) and found a non-monotonic relationship with top-1 accuracy. Performance peaks at a feature dimension of 512, suggesting that increased memory capacity generally facilitates the learning and storage of richer context--object associations. However, excessively large memory capacity may impair contextual reasoning, likely due to overfitting and reduced generalization of learned contextual knowledge.



To probe the representations learned by the external memory, we visualized the pairwise Kullback Leibler (KL) divergence of attention scores across memory slots for object categories in COCO-VOC (\textbf{Fig.~\ref{fig:fig3analysis}E}). Each matrix entry quantifies the divergence between the memory-slot attention distributions used to retrieve contextual information for a pair of object categories (see \textbf{Algo.~\ref{algo:kld}} for details). Darker values indicate greater similarity in the attended memory slots, reflecting a higher likelihood of shared contextual associations between object classes.
Across supercategories—including vehicles, animals, and indoor objects, we observe clusters of contextually related object pairs. For example, televisions and potted plants, despite limited visual similarity, exhibit strong overlap in attended memory slots, consistent with their shared indoor context. These patterns suggest that SeCo’s external memory encodes meaningful object–context associations that go beyond low-level visual similarity.\\



\noindent \textbf{Distinct target and context representations enable relational reasoning.}
SeCo was trained with two separate encoders, $E_t(\cdot)$ and $E_c(\cdot)$, for the target and context streams, respectively. We also evaluated a variant with weight-sharing between the encoders (\textbf{Fig.~\ref{fig:fig3analysis}B, Shared Encoder}). This shared-encoder baseline achieved lower top-1 accuracy than the full SeCo model, indicating that using identical features for both target and context streams is insufficient for effective context reasoning. Separating the encoders allows the model to extract complementary representations for targets and their context, which facilitates the integration of relational information necessary for accurate inference.\\

\noindent \textbf{Variance and covariance losses are critical for stable pre-training.}
SeCo has a joint loss of MSE loss, covariance loss, and variance loss. Here, we remove one loss at a time to analyze its effectiveness on pretraining. We report top-1 accuracy on the in-domain test set of COCO-OCD in \textbf{Fig.~\ref{fig:memory_slots}B}. The result demonstrates that without variance loss, SeCo reached information collapse, aligning with the trend in VICReg \cite{vicreg}. Without covariance loss, performance drops 2\% in accuracy. Different from the observations made in VICReg \cite{vicreg}, without MSE loss, SeCo manages to achieve $41.72\%$ in top-1 accuracy without collapses. One possible reason is that starting from weights obtained on ImageNet \cite{imagenet}, the encoder has captured useful visual features. Thus, adding information regularization during pre-training can avoid collapse even without enforcing association between contexts and targets.



\subsection{SeCo excels in object priming task}\label{sec:results-objectpriming}

Human vision is characterized by a remarkable ability to recognize spatial relationships and contextual relevance, enabling intuitive expectations about where objects should appear within a scene based on surrounding cues (\textbf{Fig.~\ref{fig:fig1intro}B2}). To test whether SSL models exhibit a similar capacity for context-aware object placements, we designed an object priming task for both humans and AI models.

We curated the \emph{Human Object Priming} (HOP) dataset, which comprises semantically meaningful image–target object pairs drawn from the COCO-Stuff test set \cite{cocostuff} (see \textbf{Methods}). For each image–target object pair, human priming maps were collected from participants on Amazon Mechanical Turk, who clicked contextually appropriate locations within a scene to place the target object (\textbf{Fig.~\ref{fig:fig7priming}A1}; \textbf{Methods}). To minimize crowding effects, target objects were removed from the original scenes. Aggregating these mouse-click responses yielded human object priming maps (\textbf{Fig.~\ref{fig:fig7priming}A2}; \textbf{Methods}). In parallel, AI models performed the same task on the same image–object pairs to generate machine priming maps (\textbf{Algo.~\ref{algo:priming}} and \textbf{Methods}).

Human and model priming maps were compared using a similarity metric, referred to as the ``alignment score'' (\textbf{Methods}). SeCo achieved the highest alignment with human priming maps among all baselines, indicating superior ability to predict contextually relevant object placement (\textbf{Fig.~\ref{fig:fig7priming}B}). Qualitative visualizations further illustrate these differences (\textbf{Fig.~\ref{fig:fig7priming}C}): while baseline models often produced diffuse or context-insensitive priming maps, SeCo consistently identified semantically plausible object locations. Importantly, none of the models were trained or fine-tuned using human priming data, highlighting SeCo’s remarkable capacity to transfer learned contextual associations to infer semantically appropriate, target-relevant locations within complex scenes.



\section{Discussion}

We examined how humans acquire and deploy contextual priors to reason about objects embedded in naturalistic scenes and introduced a biologically inspired computational model, SeCo (Self-supervised learning for Context reasoning, \textbf{Fig.\ref{fig:fig2arch}A}), that captures essential aspects of this capability. Across psychophysics experiments (\textbf{Fig.\ref{fig:fig1intro-human}A-D}), participants rapidly learned contextual rules from short, unlabelled videos and successfully inferred the identity of hidden objects in a “lift-the-flap” task. These findings demonstrate that humans can rapidly internalize environmental statistics and apply them to guide inference under uncertainty.

Strikingly, participants in the self-supervised condition—who received no labels or feedback—performed competitively relative to those in the supervised condition (\textbf{Fig.\ref{fig:fig1intro-human}E}). This suggests that explicit instruction is not necessary for acquiring structured contextual knowledge. Nevertheless, supervision modulated the efficiency of reasoning: reaction times were shorter in the supervised group (\textbf{Fig.\ref{fig:fig1intro-human}F}), indicating that labeled guidance may accelerate access to contextual priors. Across both learning regimes, correct trials were associated with faster responses than incorrect trials (\textbf{Fig.\ref{fig:fig1intro-human}G}), consistent with successful retrieval of coherent contextual representations rather than guess-based decision-making.

Contextual reasoning performance improved with longer training durations and curriculum-style progression (\textbf{Fig.\ref{fig:figsonsite}}), suggesting that gradual exposure and structured experience enhance the formation of contextual priors. Because humans are already highly familiar with everyday objects and their real-world associations, we curated a novel object dataset, named as FRINE dataset, to minimize the influence of prior semantic knowledge (\textbf{Fig.\ref{fig:fig1intro}D1-D4}). Specifically, we introduced four unfamiliar objects (“Fribbles”) per rule, each systematically replacing a distinct anchor object from daily environments. Within each rule, these four anchor objects defined a coherent contextual structure, and the associations were expressed through three complementary context types: global scene-level context (e.g., room-level regularities), local co-occurrence relationships, and spatial clustering patterns. This design allowed us to simulate structured yet controlled contextual environments while preventing reliance on pre-existing object knowledge. Across all three rules and all context association types, humans successfully learned and generalized the underlying structure (\textbf{Fig.\ref{fig:fig2arch}E,F}). Generalization across variations in spatial layout, context area, and resolution (\textbf{Fig.\ref{fig:fig4cond}A,B,C}) further indicates that contextual knowledge is represented at an abstract level rather than tied to specific visual configurations. Together, these behavioral findings suggest that humans learn flexible, transformation-invariant representations of object–context structure that support rapid inference even in the absence of direct object visibility.

SeCo captures these human behaviors through a computational architecture inspired by dual visual processing pathways \cite{ungerleider1982two,grill2004human,epstein1998cortical,auger2012retrosplenial,epstein2008parahippocampal} and hippocampal memory systems \cite{squire2004medial,eichenbaum2000cortical} (\textbf{Fig.\ref{fig:fig2arch}A}). Separate encoders process candidate target regions and surrounding context, loosely paralleling ventral pathways for object identity and dorsal or scene-selective regions for spatial layout. A learnable external memory module stores latent contextual associations and retrieves them when only contextual cues are available. In the lift-the-flap task, SeCo performs inference by retrieving context-linked representations and regressing toward the likely target identity, analogous to hippocampal pattern completion. 

Importantly, most existing computational studies of contextual reasoning rely heavily on supervised learning with large-scale annotated datasets to model object co-occurrence and scene structure \cite{bar2006top,bar2004visual,bar2003cortical,vo2021meaning,palmer1975effects,bar1996spatial,auckland2007nontarget, davenport2004scene,whenpigsfly,puttingincontext}. Even recent self-supervised learning approaches primarily focus on learning object-centric representations from isolated instances, optimizing invariances within single-object crops or global image embeddings \cite{contextencoder,simclr,simsiam,dino,barlowtwins,vicreg,xie2021unsupervised}. As a result, these models excel at recognizing what an object is but are not explicitly designed to learn how objects relate to one another within scenes (\textbf{Fig.\ref{fig:fig2arch}D, Fig.\ref{fig:fig4cond}D, Fig.\ref{fig:fig3analysis}A1-A4, Fig.\ref{fig:figs-glc}, Fig.\ref{fig:figs-conditions}}). In contrast, SeCo is explicitly structured to learn context–object associations without category labels, embedding relational structure within an external memory system that can be queried for inference. By shifting the focus from object identity learning to structured relational learning, SeCo provides a computational framework for modeling how contextual priors can emerge from natural visual experience with minimal supervision.

The external memory structure in SeCo reveals interpretable clusters of contextually related objects (\textbf{Fig.\ref{fig:fig3analysis}DE, \textbf{Fig.\ref{fig:memory_slots}}}), suggesting that contextual knowledge emerges as structured latent organization rather than correlated visual similarities. This parallels evidence from cognitive neuroscience proposing that semantic memory encodes relational structure linking objects within shared environments \cite{bar2004visual,mcclelland1995there,marr1971simple}. The ability of SeCo to retrieve context-conditioned representations demonstrates that contextual reasoning can emerge from self-supervised learning when models are explicitly equipped with mechanisms for structured memory storage and retrieval.

Beyond average task accuracy, it is important to examine the nature of generalization. Humans successfully transferred learned contextual rules across variations in scene appearance and spatial configuration, indicating that contextual knowledge is not tied to superficial image statistics. Similarly, SeCo generalized beyond training layouts (\textbf{Fig.\ref{fig:fig2arch}D, and Fig.\ref{fig:fig3analysis}A2,A4}), suggesting that its memory module encodes relational structure rather than memorizing individual visual exemplars. In an object priming task where both humans and machines are asked to place the target object in a contextually relevant region in the scene, SeCo’s predicted placements closely matched human choices (\textbf{Fig.\ref{fig:fig7priming}}), further supporting the idea that both humans and the model learn probabilistic priors governing object-context compatibility.

Even when SeCo approximates human contextual reasoning (\textbf{Fig.\ref{fig:fig4cond}D,E} and \textbf{Fig.\ref{fig:fig7priming}B}), the mechanisms may differ in important ways. First, the current model assumes explicit candidate region identification, whereas humans may dynamically allocate attention and integrate context over sequential fixations \cite{zhang2018finding,wang2025gazing,ding2022efficient,zhang2022look}. Second, the memory module in SeCo is implemented as a static external store; biological memory systems likely involve recurrent dynamics, consolidation processes, and interaction with broader cortical networks \cite{han2024flow,sikarwar2023decoding,wang2023object,douglas2007recurrent}. Third, learning in SeCo occurs through gradient-based optimization over curated training images, whereas humans acquire contextual knowledge through lifelong multimodal video experience with rich sensory feedback \cite{cai2025learning,zhang2025peering,jia2025seeing}. Fourth, the current experiments focus on predefined contextual rules; natural environments exhibit far more complex and hierarchical statistical dependencies \cite{wang2024unsupervised}. Finally, SeCo does not yet incorporate high-level semantic reasoning, causal understanding, or world knowledge that may further constrain human contextual inference (e.g., physical plausibility \cite{ullman2017mind} or social conventions \cite{lin2025make,piriyajitakonkij2025grunts}).

These limitations suggest several promising directions. Future work could incorporate active visual exploration, allowing models to gather contextual information sequentially. Integrating multimodal cues, such as language or action affordances, may more closely approximate real-world learning. Modeling developmental trajectories, such as how contextual priors emerge over time, could further bridge computational models and developmental psychology literature. Finally, incorporating adaptive memory updating and cross-task transfer may clarify how contextual reasoning interacts with broader cognitive systems.

The present findings advance our understanding of how contextual knowledge is acquired and applied. Rather than viewing context as a post hoc modulation of object recognition, our results suggest that contextual structure is learned as a core component of scene understanding and can guide inference even when direct sensory evidence is absent. SeCo demonstrates that self-supervised learning, when coupled with structured memory mechanisms, can approximate this human capability. Together, our behavioral and computational results support the idea that contextual scene understanding emerges from learning the statistical structure that binds objects within their environments, enabling invariant, rapid, generalizable reasoning about what belongs and what does not within a given scene. In doing so, we move closer to understanding how humans and machines learn to see the elephant in the room.

\section{Methods}



\subsection{Human Psychophysics Experiments}

To evaluate human capacity for learning contextual associations and reasoning from context, we conducted three psychophysics experiments. In the first experiment, participants learned novel contextual rules and applied them to infer the identity of a hidden object in synthetic images, a task we refer to as the \textit{lift-the-flap} task. We systematically varied experimental parameters, such as the size, resolution, and spatial configuration of contextual cues, to examine the robustness and invariance of context-based reasoning. In the second experiment, we extended the training duration and introduced a structured curriculum to further enhance the ability to acquire and apply novel context-object associations. The third experiment used real-world scenes in an object priming task, where participants placed a given object in a contextually appropriate location within the scene. Together, these experiments assess human ability to learn contextual rules and apply them to infer what an object is and where it belongs, spanning tasks from simple synthetic setups to complex naturalistic environments. All participants had normal or corrected-to-normal vision, provided written informed consent, and were compensated at a rate of 20 SGD per hour. Experiment durations varied: the first and the second lasted approximately 40-60 minutes, and the third lasted approximately 6 minutes. All studies were approved by our Institutional Review Board.


\subsubsection{Learning to Reason from Novel Context Rules}\label{sec:method:frine}

\noindent \textbf{Fribble in the Scene (FRINE) Dataset.}
Humans accumulate years of experience navigating complex scenes in daily life. To assess how effectively they can learn and apply contextual associations independent of prior knowledge, we introduced novel objects and constructed new contextual associations between these unfamiliar objects and familiar household environments. Specifically, we developed the Fribble in the Scene (FRINE) dataset using novel object families from the NOD dataset \cite{singh2023learning}. These fribbles, characterized by distinct body structures and appendages, are unfamiliar to humans. We synthesized realistic indoor scenes using the Unity-based VirtualHome environment \cite{puig2018virtualhome}, which contains seven apartments and five types of furnished rooms such as kitchens and bedrooms. By embedding fribbles into these environments, we ensured that the contextual associations encountered by participants were both ecologically valid and entirely novel.\\

\noindent \textbf{Context associations.} To study familiar contextual associations with novel objects, we replaced common household items in VirtualHome with fribbles. Specifically, we selected eight VirtualHome objects (\textbf{Fig.~\ref{fig:fig1intro}D}): \textit{keyboard}, \textit{mouse}, \textit{cake}, \textit{cup}, \textit{remote}, \textit{microwave}, \textit{toothbrush}, \textit{pillow}, and \textit{knife}, based on the types of contextual associations they exhibit (\textbf{Fig.~\ref{fig:fig1intro}B}, \textit{right}). First, \textit{global association} refers to objects that appear exclusively in a specific place without co-occurrence of other object categories; for example, toothbrushes appear only in bathrooms and microwaves only in kitchens. Second, \textit{local association} refers to objects that either appear in multiple locations or co-occur with other types; for example, keyboards always appear with computer mice, but can be located in the common room, study, or bedroom. Third, the \textit{crowding effect} describes cases where multiple instances of the same object appear together, such as eggs grouped in dozens or knives arranged in a set.\\

\noindent \textbf{Context rules.} To curate context rules, four families of fribbles in the NOD dataset (\textit{Fa1}, \textit{Fb1}, \textit{Fb3}, \textit{Fc1}) were to match with four out of the selected eight household objects (\textbf{Fig.~\ref{fig:fig1intro}D4}). Consequently, there were $A_4^8=1680$ possible combinations of replacement choices, which is impractical to handle. Therefore, we manually curated three combinations (in the format of the fribble-VirtualHome object to be replaced), where each specific combination is considered as a novel context rule: (1) Rule 1: Fa1-microwave, Fb1-mouse, Fb3-knife, Fc1-cup; (2) Rule 2: Fa1-pillow, Fb1-keyboard, Fb3-yoothbrush, Fc1-mouse; (3) Rule 3: Fa1-mouse, Fb1-cup, Fb3-pillow, Fc1-cake.\\



\noindent \textbf{Object-centric videos.} 
Humans perceive the world as a continuous stream of visual input in complex scenes \cite{singh2023learning}. To loosely approximate this visual experience, we generate video clips in which the fribble object of interest remains centered in each frame while the camera rotates around it. 
In the NOD dataset \cite{singh2023learning}, each fribble family consists of five object instances (e.g., the Fa1 family includes Fa1\_1231, Fa1\_1322, Fa1\_2333, Fa1\_3123, and Fa1\_3212). To curate video clips, we randomly select one instance from each family and replace the corresponding household object in the VirtualHome environment, following the specified context rule. A camera is configured to rotate around the fribble while maintaining a fixed attitude angle, capturing a maximum of 20 fixed viewpoints per clip. These viewpoints are spaced by 18 degrees in azimuth, ranging from 0 to 342 degrees, with a constant radius of 1.5 meters. 
To enrich the range of object views, we vary the camera's elevation angle across four levels: 0, 20, 40, and 60 degrees, generating multiple videos for the same object. Unity’s built-in ray casting and collision checking libraries are used to ensure that the camera does not collide with or get occluded by surrounding objects. 
Videos are recorded at 1 frame per second with a resolution of 1280$\times$1024 pixels, subtending $24.4 \times 19.6$
degrees of visual angle. Any frames affected by occlusions or collisions are skipped, resulting in video durations ranging from 10 to 20 seconds.\\

\label{sec:method:lor}

\noindent \textbf{Training phase of learning to reason.}
We recruited 160 human participants using Amazon Mechanical Turk (MTurk), with subjects’ informed consent. These participants were drawn from a pool of “master workers” who had completed at least 100 approved Human Intelligence Tasks (HITs) with a 95\% approval rate. In the training phase, each subject was randomly assigned to one of three context rules.
Each subject viewed 40 training videos, with 10 videos sampled per Fribble family. To assess whether participants could learn the context rules without explicit supervision, we introduced two learning modes: 
\begin{itemize}
    \item \textbf{supervised learning (SUP)}, where red bounding boxes highlighting the target object and its label (e.g., ``fa1'') were provided.
    \item  \textbf{self-supervised learning (SSL)}, where only the bounding boxes were shown without any labels. 
\end{itemize}

Subjects assigned to the same context rule and learning mode viewed the same 40 training videos. The videos were randomly shuffled and presented only once, with each lasting between 10 and 20 seconds. Video resolution was downsampled to 640$\times$512 pixels subtending $12.8 \times 10.1$
degrees in visual angles to minimize internet delays (\textbf{Fig.~\ref{fig:figsscreenshots}A1}). 

To ensure data quality and participant attentiveness, we included a memorization test with a 50\% probability after each video 
(\textbf{Fig.~\ref{fig:fig1intro}D1}, \textbf{Fig.~\ref{fig:figsscreenshots}A2}). In each test, subjects were shown a visual stimulus and asked whether they had seen the exact same object in the preceding video. These stimuli were sampled from one of the following three categories:
\begin{itemize}
\item (1) \textbf{VirtualHome-Present}: A positive sample cropped from the last frame of the current training video.
\item (2) \textbf{VirtualHome-Absent}: A negative sample cropped from an object in a different apartment, ensuring it was not present in the current video.
\item (3) \textbf{ImageNet-Absent}: A negative sample randomly drawn from ImageNet \cite{imagenet}, guaranteed to be absent from all training videos.
\end{itemize}
Each category was sampled with equal probability (33.3\%), and participants were asked: ``Did you see the EXACT SAME object instance in the last video?'' Subjects made a binary decision. To pass the quality control, participants were required to make fewer than six errors across the VirtualHome-Present and VirtualHome-Absent tests, and fewer than two errors on ImageNet-Absent tests.
The memory recall error rates are reported in \textbf{Fig.~\ref{fig:figs-memorization}A1, B1}. 
Only data from participants who passed the memorization tests were included in the final analysis. \\



\noindent \textbf{Test Phase on Lift-the-Flap: What is the target object?}
Following the training phase, participants entered the test phase involving a lift-the-flap task designed to probe the ``what'' aspect of context reasoning (\textbf{Fig.~\ref{fig:fig1intro}C1}). Each participant was randomly assigned a unique test set consisting of 40 test videos. In this task, human subjects were required to rely solely on contextual information to infer the class identity of a hidden target object in each video. The target object was obscured by a black patch, and the video could be viewed an unlimited number of times. All videos were presented at a resolution of 640$\times$512 pixels, subtending $12.8 \times 10.1$ degrees of visual angle, and had a duration ranging from 10 to 20 seconds (\textbf{Fig.~\ref{fig:figsscreenshots}B1}).\\

\noindent \textbf{Context manipulation.}
To evaluate the robustness of human reasoning under altered visual contexts, we introduced four types of context manipulations in the test videos:
\begin{itemize}
\item (1) \textbf{Normal context}: No manipulation applied.
\item (2) \textbf{Blur context}: Gaussian blur applied to the context (i.e., all video regions excluding the black-patched target area), with one of five kernel sizes ($\sigma=$ 2, 4, 8, 16, 32 pixels for frame size = 1024$\times$1280 pixels).
\item (3) \textbf{Context area}: Context was removed according to five different context-object (CO) ratios (128, 16, 8, 4, 2), where the CO ratio refers to the proportion of image area excluding the target object, relative to the size of the target object.
\item (4) \textbf{Jigsaw context}: Context was scrambled into non-overlapping pieces of size 3$\times$3, 5$\times$5, or 7$\times$7 and then shuffled.
\end{itemize}

Each manipulation was applied only to the surrounding context, leaving the target region untouched. For the CO=128 condition in the context area manipulation and the 7$\times$7 jigsaw condition, we excluded cases where the fribble object occupied more than 0.5\% of the image area. 
Each test video was randomly assigned to one of the four context manipulation types (normal, blur, area, jigsaw) with equal probability (25\%). If a video was assigned to blur, area, or jigsaw, it was then randomly assigned to one of its sub-conditions with a probability of 20\%, 20\%, or 33.3\%, respectively.

To ensure data quality and participant attention during the test phase, we embedded three dummy checks. Specifically, we selected one image each of a dog, cat, and fish from ImageNet \cite{imagenet}. These images were presented after the 10th, 20th, and 30th test videos, with the question ``Is this object a \textit{[LABEL]}?'', where ``\textit{[LABEL]}'' was randomly substituted with a label, either ``dog'', ``cat'', or ``fish''. Participants were required to make a binary judgment on whether the shown object matched the given label (\textbf{Fig.~\ref{fig:figsscreenshots}B2}). Only those who correctly answered all three dummy tests passed the quality check.
Out of the 160 participants, 108 passed the quality controls for both the training and test phases, resulting in 4320 valid trials. Each context sub-condition within each learning mode included an average of at least 85 trials.

\subsubsection{Learning to Reason with Longer Training and Better Curriculum}
In education, students follow a structured curriculum where each new concept builds on and reinforces prior knowledge. Thoughtful curriculum design is essential for improving learning in humans \cite{singh2023learning, tee2023integrating}. 
To investigate whether human reasoning from context can be further improved, we extended the training duration and implemented a structured curriculum in a controlled lab setting, free from distractions to ensure participants remained attentive throughout the experiment. 
We recruited a total of 20 participants. Each was assigned to learn only Context Rule 1 and randomly placed into either a supervised or self-supervised learning mode. All subjects viewed the same set of 40 training videos corresponding to Context Rule 1 but each video was shown multiple times according to the structured curriculum design: (1) in the first stage, 10 videos from each fribble category were shown sequentially; (2) in the second stage, 5 videos from each category were interleaved; (3) in the final stage, 10 videos were presented from the two most difficult fribble categories identified in earlier experiments (Fb3 and Fc1). In total, each participant viewed 80 training videos, each with a resolution of 640$\times$512 pixels subtending $12.8 \times 10.1$ degrees in visual angles and their durations ranging from 10 to 20 seconds. As in the previous online experiments, memorization tests were included during training.

Participants then proceeded to the same testing phase as in the previous online experiments, performing the lift-the-flap task. For consistency, each subject was randomly assigned a unique test set of 40 videos drawn from the online experiment pool. All four context manipulations, and the dummy checks during testing, were retained. A total of 15 out of 20 participants passed the quality control criteria, resulting in 600 valid trials, with at least 75 trials per context manipulation condition per learning mode on average.


\subsubsection{Lift-the-Flap: What Is the Target Object behind the Flap?}

\textbf{COCO-OCD and COCO-VOC Datasets}

To further evaluate the reasoning capabilities of AI models, we considered the real-world scenario with inherently complex scenes featuring diverse objects, environments, and unpredictable spatial relationships. We specifically considered one base natural scene dataset COCO-Stuff \cite{cocostuff}, which originally contains 160K natural images from MSCOCO \cite{mscoco} with 80 thing classes and 91 stuff classes in total. Importantly, this dataset captures complex relationships between multiple objects and carries rich contextual information. To evaluate whether the learned contextual knowledge
from AI models can generalize well in out-of-domain
settings, we considered two base datasets: (1) Out-of-Context Dataset (OCD) \cite{whenpigsfly}, which originally contains 15,773 synthetic test images of indoor scenes with 36 classes under 6 different contextual conditions. In our work, we only consider the normal context condition with 2,309 test images. (2) PASCAL-VOC07 Dataset \cite{voc07}, which originally contains 9,963 images of realistic scenes with a total of 20 object classes. We designed two custom datasets, curated from COCO-Stuff, OCD and PASCAL-VOC07 as follows (\textbf{Fig.~\ref{fig:dataset}}).

\noindent \textbf{COCO-OCD}. The custom dataset includes one training set derived from COCO-Stuff and two test sets derived from COCO-Stuff and OCD. The training and test images are from 15 overlapping classes shared between the two base datasets: wine glass, cup, knife, bowl, apple, cake, mouse, remote, keyboard, cell phone, microwave, book, toothbrush, pillow, and towel. We refer to test images from COCO-Stuff as in-domain COCO-OCD, and test images from OCD as out-of-domain COCO-OCD.

\noindent \textbf{COCO-VOC}. The custom dataset includes one training set derived from COCO-Stuff and two test sets derived from  COCO-Stuff and PSACAL-VOC07. The training and test images are from 20 overlapping classes shared between the two base datasets, in a hierarchy of \textit{superclass} and subclass: (1) \textit{Person}: person;
(2)\textit{Animal}: bird, cat, cow, dog, horse, sheep;
(3) \textit{Vehicle}: aeroplane, bicycle, boat, bus, car, motorbike, train;
(4) \textit{Indoor}: bottle, chair, dining table, potted plant, sofa, tv/monitor.
We refer to test images from COCO-Stuff as in-domain COCO-VOC, and test images from PASCAL-VOC07 as out-of-domain COCO-VOC.

\subsubsection{Object Priming: Where to Put the Target Object in the Scene?}

To further evaluate the reasoning capabilities of human subjects, we introduce the object priming task to address the problem of ``where" in context reasoning (\textbf{Fig.~\ref{fig:fig1intro}C2}) in complex real-world scenes featuring diverse objects, environments, and regular spatial relationships. 
In this task, human subjects are given an image and a target object as inputs and have to predict contextually correct locations for placing the target object. 

To curate a dataset of semantically meaningful image–object pairs for the object priming task, we sampled images from the COCO-OCD test set that contained at least three object classes from the 15 selected categories. For each image, we manually assigned a target class—corresponding to the object to be placed—by selecting a semantically appropriate category from the same 15 classes, while ensuring that the target class was not already present in the image. This procedure resulted in 864 semantically meaningful image–object pairs drawn from 206 unique natural images. We refer to this dataset as the \emph{Human Object Priming} (HOP) dataset. 

We recruited a total of 437 human participants via Amazon Mechanical Turk (AMT). For each participant, we randomly sampled 20 image--object pairs and presented each of them to human subjects with an image of size $800 \times 800$ pixels, subtending $15.56^\circ \times 15.56^\circ$ of visual angle, together with the question ``Where would you put this \emph{[obj]}?'', where \emph{[obj]} denotes the sampled target object. Participants were instructed to make 10 non-repeated mouse clicks on semantically relevant regions of the image (\textbf{Fig.~\ref{fig:figsscreenshots}C}). Each image--object pair was completed by 3--7 participants. To ensure equal coverage across all image--object pairs, we randomly subsampled click data from 3 unique participants per pair (10 clicks each; 30 clicks total). These 30 click coordinates were consolidated to generate human priming maps.

Specifically, we first divided each $800 \times 800$ stimulus image into a $32 \times 32$ grid, yielding 1{,}024 cells of size $25 \times 25$ pixels. Mouse clicks were aggregated within each grid cell such that the pixel intensity reflected the number of clicks. The resulting $32 \times 32$ priming map was then smoothed using an $11 \times 11$ Gaussian filter, resized to $224 \times 224$, and min--max normalized to produce the final human priming maps.

\subsection{Models}
\subsubsection{SeCo}
\label{section:SeCo}

We propose a Self-Supervised Learning Method for Context Reasoning (SeCo) which learns associations between objects and their contexts in natural images (\textbf{Fig.~\ref{fig:fig2arch}A)}. SeCo consists of three components:
(a) object discovery module, (b) two-stream visual processor, and (c) external memory. 
First, the target discovery module uses unsupervised region proposal methods to locate potential objects of interest on a full image $I_f$. 
Each region proposal together with the full image $I_f$ is subsequently converted to pairs of target images $I_t$ and context images $I_c$.
Second, the two-stream visual processor consists of two independent convolutional neural network (CNN) encoders and projectors, extracting information from $I_t$ and $I_c$, respectively.
Third, SeCo employs a trainable external memory
to store knowledge priors about contextual associations learned during training phase. 
Features from $I_c$ serve as queries to retrieve context-relevant prior knowledge from the external memory with an attention mechanism. The retrieved information provides the complementary signal to the context stream and gets compared with the target features from $I_t$ of the object stream to maximize the agreement between the stored prior knowledge and the context-relevant object in the embedding space (see \textbf{Algo.~\ref{algo:seco}} 
for the PyTorch-style pseudocode of SeCo's pre-training algorithm).\\

\begin{algorithm}[t]
\small
\SetAlgoLined
    \PyComment{Ec, Et: context and target encoders} \\
    \PyComment{pc, pt: context and target projectors} \\
    \PyComment{M: external memory shaped in K-by-H} \\
    \PyComment{pk: key projection of external memory}\\
    \PyComment{mse: mean square error loss} \\
    \PyComment{var\_loss: variance loss} \\
    \PyComment{cov\_loss: covariance loss} \\
    \PyComment{alpha, beta, gamma: weightage of each loss component} \\    

    \PyComment{load a batch of N images} \\ 
    \PyCode{for x in loader:} \\
    
    \Indp   
        \PyComment{randomly augmented target and context} \\
        \PyCode{t, c = augment(x)}\\ 
        
        \PyComment{encode and project context, target stream} \\
        \PyCode{hc, ht = Ec(x), Et(x)} \PyComment{N x D} \\
        \PyCode{sc, st = pc(hc), pt(ht)} \PyComment{N x H} \\
        \PyComment{compute keys of memory} \\
        \PyCode{m = pk(M)} \PyComment{K x H} \\
        \PyComment{retrieve memory} \\
        \PyCode{p = softmax(dot(sc, m))/sqrt(H)} \PyComment{N x K} \\
        \PyCode{sc = p * M} \PyComment{N x H} \\
        \PyComment{calculate loss and update} \\
        \PyCode{loss = alpha * mse(sc,st) + beta * (var\_loss(sc) +  var\_loss(st)) / 2  + gamma * (cov\_loss(sc)+  cov\_loss(st))}\\
        \PyCode{loss.backward()}
        
    \Indm 
\caption{PyTorch-style pseudocode for SeCo}
\label{algo:seco}
\end{algorithm}

\noindent\textbf{Context-Object Pair Discovery}. Objects play an important role in context reasoning \cite{draschkow2017scene}. To learn 
object-object and object-context associations, 
we propose a context-object pair discovery module to exploit regions containing objects of interest.      
We adopt the selective search algorithm \cite{selectivesearch} to generate regions of interest (RoI) that potentially contain objects. It is worth noting that selective search is an unsupervised learning algorithm. It performs heuristic searches on hundreds of anchor boxes and proposes RoIs by hierarchically grouping similar regions based on color, texture, size, and shape compatibility. 
To reduce false positives among many RoIs, we
filter out resultant regions according to their area ratio (with a maximum of 0.1) and aspect ratio (within 0.2 and 5). Moreover, we merge RoIs with heavy overlaps by setting the threshold of IoU (intersection over union) as 0.3. 
For each selected RoI, we generate a pair of target images $I_t$ and context image $I_c$. $I_t$ is cropped out of full image $I_f$. The entire image with the RoI blacked out with zeros forms the context image $I_c$.\\

\noindent\textbf{Feature Extraction with CNN}. Due to eccentricity dependence, human vision has the highest acuity at the fovea and the resolution drops sharply in the periphery with increasing eccentricity \cite{gupta2021visual,freeman2011metamers}.
For example, while we are fixating on the mug on the table, the mug is often perceived in high resolution while the context gist of the kitchen scene is processed at low resolution in the periphery. Taking inspiration from this observation, we propose a two-stream visual processor, with one object stream dedicated to encoding the target image $I_t$ and the other context stream dedicated to encoding the context image $I_c$. The encoded representations are denoted as $h_c=E_c(I_c)$ and $h_t=E_t(I_t)$, where $E_t(\cdot)$ and $E_c(\cdot)$ are target and context encoders and $h_t$ and $h_c \in \mathbb R^{D}$. 
Of note, we do not enforce weight sharing between the encoders. We demonstrate the benefit of this approach in the model analysis.\\

\noindent\textbf{Training With An External Memory}. As suggested by cognitive and neuroscience works \cite{puttingincontext,riesenhuber1999hierarchical,thorpe1996speed}, context processing often happens very fast in the brain. The perceived scene gist serves as a query to retrieve prior knowledge from the semantic memory to modulate object recognition in a top-down manner. To mimic this underlying mechanism of context modulation,
we introduce an external memory with trainable parameters, accumulating prior knowledge of 
contextual associations.
Different from the well-established cross-attention mechanism \cite{vaswani2017attention}, the objective of our external memory focuses on dynamically retrieving and updating the prior knowledge.

We define the external memory as a 2D matrix with trainable parameters, which consists of $K$ memory slots of $H$ dimension, denoted as $M = \{m_1, ..., m_K \}, M \in \mathbb R^{H\times K}$. Each memory slot is associated with a key, where $P_k(\cdot): \mathbb{R}^{H}\to\mathbb{R}^{H}$ defines the linear mapping from the memory content to the keys $P_k(M)$. The encoded representation $h_c$ from the context stream serves as queries to the external memory after a projection operation 
$P_c(\cdot):\mathbb{R}^{D}\to\mathbb{R}^{H}$. 
The retrieved prior knowledge $s_c \in \mathbb{R}^{H}$ from $M$ can then be represented as 
\begin{equation}\label{equ:scmem}
    s_c=\text{SOFTMAX}(\frac{P_c(h_c)P_k(M)^T}{\sqrt{H}})M
\end{equation}
where \text{SOFTMAX}$(\cdot)$ is the standard softmax operation.\\

\noindent\textbf{Loss Components}. To encourage $M$ to learn rich and meaningful context-object associations, we introduce three types of losses. Ideally, given only the scene gist, the retrieved prior $s_c$ from $M$ should represent useful object information related to the given context (i.e., ``what could be the target object given the scene gist" versus ``the actual object seen in the scene"). Thus, we apply a mean squared error loss ${l}_{mse}$ to maximize the agreement between $s_c$ and $h_t$. To make the vector dimension comparable, $h_t$ is projected to $s_t \in \mathbb R^{H}$ in the embedding space via the projection $P_t(\cdot)$.

As shown by previous works in non-contrastive learning \cite{vicreg,simsiam}, maximizing the agreement between two-stream visual processors alone may lead to model collapses (e.g., where 
the external memory stores and outputs trivial knowledge of all zeros, while the visual processor encodes images to representations of all zeros). In this case, $s_c$ and $s_t$ align perfectly, but the encoded object representations and content in $M$ are meaningless.

Thus, to prevent model collapses, we follow \cite{vicreg} to enforce covariance ${L}_{cov}$ and variance ${L}_{var}$ regularization on both object and context streams. ${L}_{var}$ maintains the variance of batch-wise representations, encouraging object class diversities, while
${L}_{cov}$ de-correlates channel-wise variables to diversify attributes of an embedding, i.e., maximize independent attributes to represent objects. SeCo is jointly trained with the total loss:
\begin{equation}
    {L}_{total} = \alpha{L}_{mse}(s_c,s_t) + \beta [{L}_{var}(s_c) + {L}_{var}(s_t)] 
    + \gamma [{L}_{cov}(s_c) + {L}_{cov}(s_t)]
\label{eqn:loss_eqn}
\end{equation}
where $\alpha = 25$, $\beta = 25$ and $\gamma = 1$ are hyper-parameters weighting different loss components.\\

\noindent \textbf{Data Augmentations}. Data augmentation techniques are widely used at image levels in SSL literature. We applied standard image augmentations on both $I_t$ and $I_c$, including color jitter, grayscale, horizontal flip, gaussian blur, and color normalization using the existing library PyTorch. Moreover, the random resized crop is another effective technique in SSL. However, directly applying this approach is not feasible in our case. Thus, we extended the standard approach to context-object image pairs with context-aware crops by ensuring that the relative locations among objects are preserved and the bounding box encompassing the target object is always intact and present on $I_c$ after geometric transformations.\\ 

\noindent \textbf{Model Architecture}. We use ResNet-$50$ \cite{resnet} with $D = 2048$ output units as encoders $E_c$ and $E_t$. 
We set the size of $M$ as $K\times H = 200\times8192$ and initialize $M$ by the Xavier uniform initializer \cite{xavierinit}. Following VICReg \cite{vicreg}, the projectors $P_c$, $P_t$, and $P_k$ are implemented as three fully connected layers with 8192 units per layer and ReLU activations between layers. 



\subsubsection{Baselines}\label{sec:method:baselines}
Given that ground truth labels are costly to obtain for supervised learning and that much larger datasets can be used without labels, SSL has become an emerging trend in AI. We considered seven state-of-the-art SSL methods with different types of pretext tasks: 
(1) Context Encoder\cite{contextencoder} is trained to reconstruct missing image regions based on the surrounding pixel context. Since Context Encoder was originally trained with AlexNet \cite{alexnet}, we re-implemented it with the standard ResNet-50 backbone ({\textbf{Fig.~\ref{fig:figs-contextenc}}}) for fair comparisons with SeCo and other SSL methods. 
(2) SimCLR \cite{simclr} is a contrastive learning framework that maximizes agreement between differently augmented views of the same image. 
(3) SimSiam \cite{simsiam} is a non-contrastive method that minimizes the negative cosine similarity between the prediction from one branch and the (stop-gradient) projection from the other branch. 
(4) DINO \cite{dino} is a self-distillation approach where a student network mimics a momentum-updated teacher. 
(5) VICReg \cite{vicreg} prevents collapse by explicitly regularizing embedding variance and covariance. 
(6) ORL \cite{xie2021unsupervised} takes an object-level approach that leverages unsupervised masks to align object representations. 
(7) The supervised learning baseline was trained using a classification loss with ground-truth labels of the target object.
For fair comparisons, all the methods used standard ResNet-50 backbones, with weights pre-trained on ImageNet obtained from their own public checkpoints. We used the same implementations from their original papers.


\subsubsection{Experimental Setups for Quantity-quality Analysis of Selective Search}

To investigate the role of selective search in SeCo, we compared the default model against three variants that modify the generation of target and context image pairs ($I_t, I_c$) (\textbf{Fig.~\ref{fig:fig3analysis}B}). In the \textbf{Ground Truth} variant, the context-object discovery module is replaced with human-annotated bounding boxes, ensuring that target images contain precisely localized objects. The \textbf{Random Generation} variant preserves the same number of region proposals as the default SeCo model (Selective Search) but replaces them with randomly generated coordinates, thereby testing the importance of object-centric visual signals. The \textbf{Random Cropping} variant forms the target–context pair by applying two independent random crops to the same input image, removing the explicit fovea–periphery relationship induced by the discovery module.

To further examine how the quantity and quality of region proposals influence SeCo pre-training, we systematically varied both factors. We first scored each proposal using its intersection-over-union (IoU) with ground-truth bounding boxes. We retained images from the COCO-OCD training set that contained at least 10 proposals with IoU greater than 0.3, resulting in 19.7K images. The following protocols were used to benchmark SeCo with respect to proposal quality and quantity.

\noindent
\textbf{[Quality].} We filtered proposals using an IoU threshold $\gamma$, yielding a proposal pool $\mathbf{B}_\gamma = \{ b \mid IoU(b) > \gamma \}$. To isolate proposal quality, we fixed the number of proposals per image $I^i$ and varied only $\gamma$. Specifically, we randomly selected five proposals from $\mathbf{B}^i_\gamma$ while varying $\gamma \in \{0, 0.1, 0.2, 0.3\}$. Training and evaluation followed the procedure described in \textbf{Sec.~\ref{sec:method:lor}}.

\noindent
\textbf{[Quantity].} We fixed the IoU threshold at $\gamma = 0$. For each image $I^i$, we varied the number of proposals in $\{5, 10, 20, 30\}$ by randomly sampling from $\mathbf{B}^i_{\gamma=0}$. Training and evaluation followed the same procedure described in \textbf{Sec.~\ref{sec:method:lor}}.


\subsubsection{Experimental Setups for Memory Configurations}

To examine the role of stored semantic priors in contextual reasoning, we incorporate a learnable external memory module into the SeCo architecture (see \textbf{Sec.~\ref{section:SeCo}}). This module is intended to emulate the medial temporal lobe and hippocampal systems of the brain, which encode and retrieve contextual associations acquired from exposure to natural environments.

We investigated the impact of memory configurations by systematically altering the encoding and memory components (\textbf{Fig.~\ref{fig:fig3analysis}B}). In the \textbf{No Memory} variant, we remove the external memory module to evaluate the role of stored contextual priors at inference, forcing the model to rely exclusively on the projection of the encoded context features $s_c = P_c(h_c)$ in contrast to \textbf{Equ.\ref{equ:scmem}}.
Furthermore, we varied memory capacity by adjusting the number of memory slots $K$ and the feature dimensionality $H$ (\textbf{Fig.~\ref{fig:memory_slots}A}). 
Lastly, we introduce the \textbf{Shared Encoder} variant, which enforces weight sharing between the target encoder $E_t$ and the context encoder $E_c$, to evaluate whether a unified feature representation is sufficient for relational reasoning.

To interpret the contextual associations learned by the external memory, we provide a visualization based on attention score distributions. For each sample in the COCO-VOC evaluation set, we identify the memory slot receiving the highest attention weight. These activations are aggregated into category-wise distributions to determine which slots are most frequently associated with specific object classes. To quantify the overlap in contextual knowledge across categories, we compute the pairwise Kullback–Leibler (KL) divergence between the corresponding attention distributions (see \textbf{Algo.~\ref{algo:kld}}). This analysis provides a measure of shared contextual associations among different object categories.


\begin{algorithm}[!h]
\small
\SetAlgoLined
\PyComment{Ec: context encoders} \\
\PyComment{pc: context projector} \\
\PyComment{M: external memory shaped in K-by-H} \\
\PyComment{F: frequency matrix-shaped in C-by-K} \\
\PyComment{D: pair-wise KL-divergence matrix-shaped in C-by-C} \\
\PyComment{product: cartesian product of two sets}\\
\PyComment{kld: KL-divergence function}\\

\PyCode{for x, label in loader:} \\
    
    \Indp 
    
        \PyComment{obtain erased context} \\
        \PyCode{c = erase(x)}\\ 
        \PyCode{}\\
        \PyComment{encode and project context stream} \\
        \PyCode{hc = Ec(x)} \PyComment{1 x D} \\
        \PyCode{sc = pc(hc)} \PyComment{1 x H} \\
        \PyComment{compute keys of memory} \\
        \PyCode{m = pk(M)} \PyComment{K x H} \\
        \PyCode{}\\
        \PyComment{retrieve attention score over memory slots} \\
        \PyCode{p = softmax(dot(sc, m))/sqrt(H)} \PyComment{1 x K} \\
        \PyComment{sharpen the distribution} \\
        \PyCode{top1 = p.max(0)[1]} \\
        \PyCode{F[label, top1] += 1 } \\
    
    \Indm


\PyComment{calculate pairwise KL-divergence} \\
\PyCode{for i,j in product(range(C), range(C)):} \\
    
    \Indp
    
    F[i] = (F[i] - F[i].min()) / (F[i].max() - F[i].min()) \\
    F[j] = (F[j] - F[j].min()) / (F[j].max() - F[j].min()) \\
    pi, pj = softmax(F[i]), softmax(F[j])\\
    D[i,j] = kld(pi, pj)
    
    \Indm
    
\caption{PyTorch-style pseudocode for calculating pairwise KL divergence of attention score over memory slots for object categories in COCO-VOC. }
\label{algo:kld}
\end{algorithm}

\subsubsection{Experimental Setups for Loss Component Configurations}
To disentangle the individual contributions of different information constraints to stable pre-training and contextual association, we analyze the joint training objective of SeCo. The overall loss comprises three terms: mean squared error ($L_{mse}$), variance regularization ($L_{var}$), and covariance regularization ($L_{cov}$), with their relative contributions controlled by hyperparameters $\alpha$, $\beta$, and $\gamma$ (see \textbf{Eqn.~\ref{eqn:loss_eqn}}). We introduce ablated variants that systematically remove one loss component at a time to assess its role in enabling contextual association without representational collapse. The results show that all three components are necessary for optimal performance, while removing any single term leads to degraded performance or training instability, including model collapse (\textbf{Fig.~\ref{fig:memory_slots}B}).

\subsubsection{Benchmarking AI Models}


\noindent\textbf{Training AI models for lift the flap tasks on naturalistic images}. 
SeCo and SSL baselines (\textbf{Sec.~\ref{sec:method:baselines}}) were initialized with ImageNet-pretrained weights, followed by pre-training on the COCO-OCD 
and COCO-VOC training sets respectively. During the pre-training of SeCo, we set the base learning rate to $lr=0.05*\text{batch\_size}/256$ \cite{linearscale} for COCO-OCD and $lr=0.1*\text{batch\_size}/256$ for COCO-VOC. The learning rate grows linearly from $0$ to base value during the first $10$ epochs and then decays with a cosine scheduler \cite{cosinescheduler} for the rest of epochs with a minimum value of $0.0002$. Pre-training protocols for SSL baselines follow those in their original implementations. As ImageNet-pretrained weights for ORL \cite{xie2021unsupervised}
were unavailable, we directly used weights pre-trained on the full COCO-Stuff dataset \cite{cocostuff}. For the supervised learning baseline, we directly used weights pre-trained on ImageNet \cite{imagenet}. 
After the pre-trained weights for models were obtained, we fine-tuned linear classifiers on the top of frozen encoders for each model on the COCO-OCD and COCO-VOC training sets respectively, using the cross-entropy loss. We conducted the hyperparameter search for each model to find the best learning rate for fine-tuning. The final learning rates used for fine-tuning each model are as follows, decaying with a cosine scheduler \cite{cosinescheduler} for 100 epochs: 0.1 (SeCo), 10 (ORL), 0.1 (VICReg), 0.1 (DINO), 20 (SimSiam), 0.1 (SimCLR), 10 (Context Encoder), 0.1 (Supervised Learning).\\

\noindent\textbf{Benchmarking AI models for lift-the-flap}. 
We evaluated the models trained in LoR training stage (\textbf{Sec.~\ref{sec:method:lor}}) on COCO-OCD, COCO-VOC, and FRINE datasets. 
Specifically, models trained on in-domain COCO-OCD training set were firstly tested on in-domain COCO-OCD test set, then tested on out-of-domain COCO-OCD and FRINE test set in a zero-shot manner; Similarly, models pre-trained on in-domain COCO-VOC training set were firstly tested on in-domain COCO-VOC, then tested on out-of-domain PASCAL-VOC07 test set in a zero-shot manner. Of note, the FRINE test set used for AI models contains static frame images from all the test videos viewed by human subjects. To adapt linear classifiers trained on COCO-OCD (15 classes of real-world objects) to the FRINE dataset (4 classes of novel object fribbles) in a zero-shot manner, we selected the logits corresponding to the 4 VirtualHome objects associated with the fribbles (e.g., Fa1-microwave, Fb1-mouse, Fb3-knife, Fc1-cup) from the 15 logits produced by the classifier (See \textbf{Sec.~\ref{sec:method:frine}} for detailed context rules). The selected logits, representing the categories of VirtualHome objects before replacement, were then passed through the softmax function. The fribble class corresponding to the highest probability was taken as the final prediction.\\

\noindent\textbf{Benchmarking AI models for object priming}. To predict priming maps for all the models, we converted the object priming task to a series of lift-the-flap tasks 
with the following steps: (1) we divide the context image into patches. (2) We covered a single image patch with a flap (black pixels) while the remaining patches remained intact. (3) We tested all models fine-tuned on COCO-OCD from the lift-the-flap task in (2) and recorded the predicted classification probability of the model for the given target object class in the object priming task. (4) We iterated through {\footnotesize (2)} and {\footnotesize (3)} until we exhaustively performed ``lift-the-flap" tasks over all the image patches. (5) For each image patch, we then have a classification score indicating how confidently the model would put the given target object in that patch. We consolidated all the probabilities for all the patches and generated the priming map for each model. As the model predictions were sensitive to the patch sizes, we varied the patch sizes and normalized the final priming map over all patch sizes (see \textbf{Algo.~\ref{algo:priming}} for the PyTorch-style pseudocode of object priming algorithm).

\begin{algorithm}[t]
\small
\SetAlgoLined
    \PyComment{Ec: trained context network with an encoder and a linear classifier} \\
    \PyComment{patch\_sizes: patch sizes when making erased contexts} \\

    
    \PyComment{load a batch of N images} \\ 
    \PyCode{for x, label in loader:} \\
        
        \Indp 
        
        \PyCode{maps = []}\\
        \PyCode{}\\
        \PyComment{calculate priming maps in multiple scales} \\ 
        \PyCode{for patch\_size in patch\_sizes:} \\
        
        \Indp   
        
            \PyComment{iteratively erase a patch from image }\\
            \PyCode{contexts = make\_context(x, patch\_size)} \\
            \PyCode{}\\
            \PyComment{retrieve probability w.r.t location for a given object category}\\
            \PyCode{p = softmax(Ec(x)[:,label])}\\
            \PyCode{}\\
            \PyComment{normalize so that priming maps in different scales can add up}\\
            \PyCode{p = (p - p.min()) / (p.max() - p.min())}\\
            \PyCode{}\\
            \PyComment{upsample to the size of input image}\\
            \PyCode{patch\_num = x.size[1] // patch\_size} \\
            \PyCode{p = p.view((patch\_num,patch\_num))}\\
            \PyCode{p = upsample(p)}\\
            \PyCode{maps.append(p)}\\
            
        \Indm
        
        \PyCode{}\\
        \PyComment{finalize priming maps by averaging and normalizing over different scales}\\
         \PyCode{maps = torch.stack(maps).mean(0)} \\
         \PyCode{maps = (maps - maps.min()) / (maps.max() - maps.min())}\\
    
    \Indm
    
\caption{PyTorch-style pseudocode for generating priming maps. }
\label{algo:priming}
\end{algorithm}


\subsection{Data Analysis}

\subsubsection{Behavioral Metrics}
\noindent \textbf{Top-1 Accuracy.} The lift-the-flap task is formulated as a multi-class classification problem, where the goal is to predict the identity of the hidden target object from a set of candidate categories. Top-1 accuracy is defined as the proportion of trials in which the predicted category matches the ground-truth label of the hidden target. Performance is summarized as the mean trial-wise correctness, with the standard error of the mean (SEM) quantifying uncertainty.\\

\noindent \textbf{Alignment Score.} To quantify spatial reasoning in the object priming task, we measure the agreement between human and AI model predicted priming maps. For each image–object pair, we first compute the root mean square error (RMSE) between the model-predicted and human priming maps. This RMSE is then normalized by the RMSE obtained from a chance model generating random priming maps, resulting in a normalized RMSE (nRMSE) ranging from 0 to 1. Finally, the agreement score is defined as $1 - \text{nRMSE}$, so that higher values indicate greater alignment between the model and human priming maps.\\

\noindent \textbf{Pearson Correlation.} To assess the consistency of behavioral sensitivity between humans and AI models, we compute the Pearson correlation coefficient over top-1 accuracy across different contextual manipulations, including blur, reduced area, and jigsaw contexts. This metric captures how closely models replicate human contextual reasoning patterns, with higher correlation values indicating that the model and humans share similar behavioral biases in reasoning from contextual cues.

\subsubsection{Statistical Analyses}

\noindent \textbf{Bootstrapping and Distribution Analysis.}
In each experiment, we assessed the statistical robustness of performance metrics by generating empirical distributions of the mean top-1 accuracy using bootstrapping. Specifically, we performed resampling with replacement for 1,000 iterations to obtain a stable distribution of bootstrap means for each condition. Welch’s two-tailed t-tests were then used to evaluate differences between conditions and to compare experimental results against chance levels, with statistical significance determined using a consistent threshold of $p < 0.05$. Throughout the manuscript, we used the term ``Welch’s two-tailed t-test" to assess statistical differences between bootstrapped distributions.\\

\noindent \textbf{One-way Analysis of Variance (ANOVA).} To evaluate whether categorical experimental manipulations significantly influenced reasoning performance, we applied one-way analysis of variance (ANOVA). This test was used to compare bootstrapped mean top-1 accuracy across multiple levels of a single context manipulation family, including variations in context blur, context–object size ratios, and jigsaw permutations. The resulting $p$-value was used to determine whether at least one condition within each multi-level group differed significantly from the others. Statistical significance was determined using a consistent threshold of $p < 0.05$.

\subsection{Code and Data Availability}
All the source code and raw data are made publicly available in this submission through the following repository: \href{https://github.com/ZhangLab-DeepNeuroCogLab/SelfSupervisedContextReasoning}{GitHub Link}.

\section*{Acknowledgments}

This work was supported by the National Research Foundation, Singapore under its NRFF award NRF-NRFF15-2023-0001 and Mengmi Zhang's Startup Grant from Nanyang Technological University. 

\section*{Author Contributions}
The tasks were designed by XL, ZS, GK, and MZ.
XL, AS, and BAK were involved in collecting the data. The data were analyzed by XL, SS, and AS, under supervision by MZ. The manuscript was written by XL, SS, AS, GK, and MZ, and was approved by all the authors. 

\section*{Competing Interests}
The authors declare no competing interests.

{\small
\bibliographystyle{ieee_fullname}
\bibliography{cite}
}
\newpage
\section*{Main Figures}




\newpage


\newpage

\begin{figure}[!ht]
\begin{center}
\includegraphics[width=15cm]{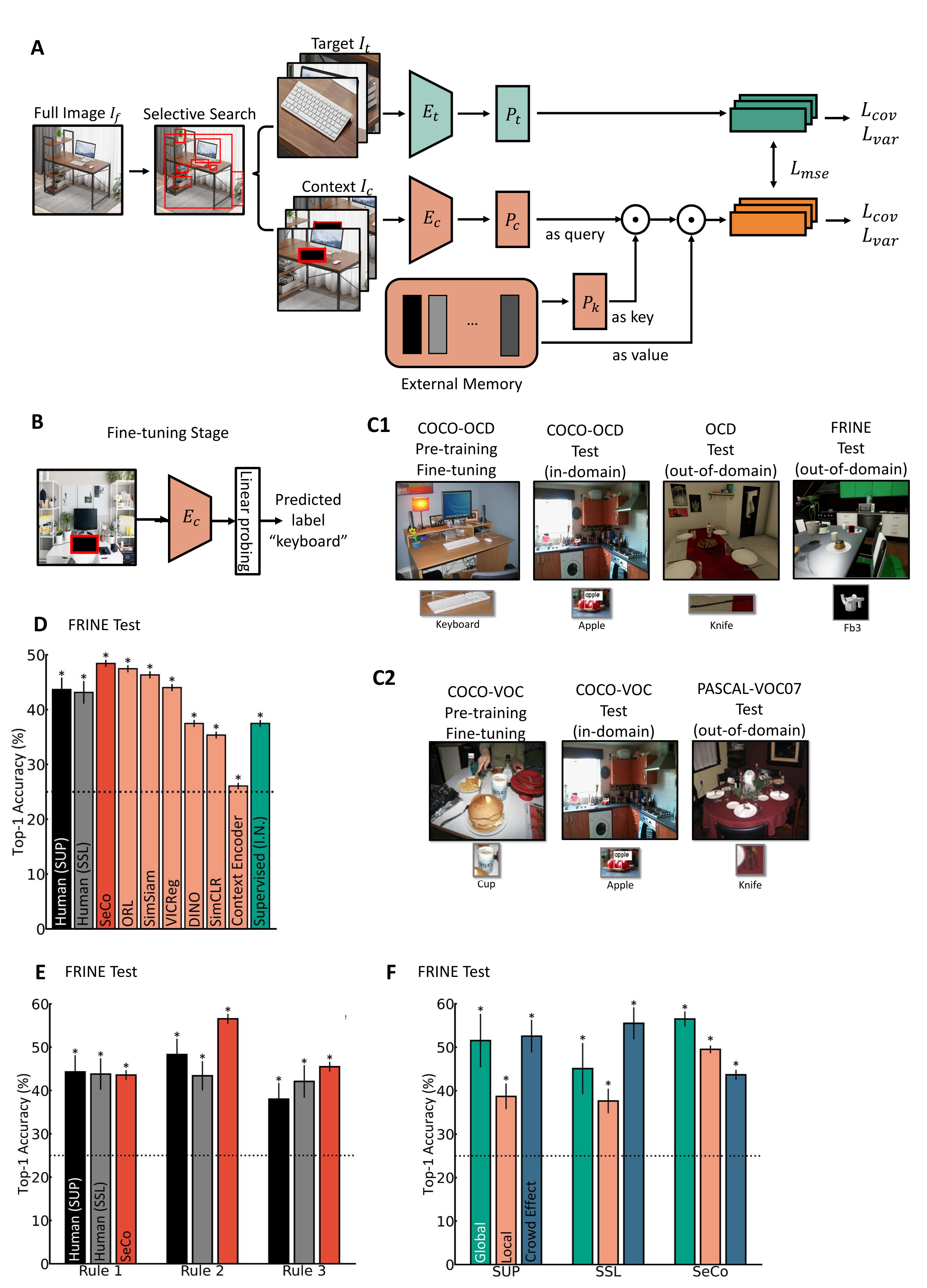}\vspace{-4mm}
\end{center}
\caption{}    
\vspace{-5mm}
\label{fig:fig2arch}
\end{figure}

\begin{figure}
  \ContinuedFloat
  \centering
  \begin{minipage}{15cm} 
    \captionsetup{singlelinecheck=off, justification=justified}
  \caption{
  \footnotesize \textbf{We propose the Self-supervised Method with External Memories for Context Reasoning, named as SeCo.}
  \textbf{A.} Model schematic of SeCo. The architecture consists of three components: a target discovery module, a two-stream visual processor (trapezoids), and an external memory module (orange squircle). During pre-training, given a full image $I_f$, SeCo employs an unsupervised method (selective search) to generate potential object proposals. These proposals are then converted into multiple context–target image pairs $(I_c, I_t)$. For each pair, SeCo processes the context and target separately using non-shared encoders ($E_c$, $E_t$) followed by non-shared projectors ($P_c$, $P_t$).
  In parallel, SeCo incorporates a trainable external memory to store contextual priors. Ideally, if $E_c$ encodes strong contextual cues, its latent representation should serve as an effective query to retrieve a relevant object representation from the external memory based on learned keys $P_k$ and values. A joint training objective is applied: the mean squared error loss $L_{\text{mse}}$ encourages alignment between the retrieved embedding from the memory (based on context) and the target object embedding, while additional regularization losses—the covariance loss $L_{\text{cov}}$ and the variance loss $L_{\text{var}}$—promote diversity and prevent collapse of the learned representations (shown as stacked rectangles). 
  \textbf{B.} During fine-tuning, SeCo is adapted to the downstream lift-the-flap task. Given a test image with a hidden object, the pre-trained frozen context encoder $E_c$ is used to extract contextual representations, followed by a fully connected layer to predict the label of the hidden object (e.g., "keyboard") behind the black patch via linear probing. 
  \textbf{C.} Datasets used for pre-training, fine-tuning, and testing AI models.  
\textbf{C1} The COCO-OCD dataset contains naturalistic images from COCO-Stuff \cite{cocostuff} overlapping with object classes in the OCD dataset \cite{whenpigsfly}.  
\textbf{C2} The COCO-VOC dataset contains COCO-Stuff images overlapping with object classes in PASCAL-VOC07 \cite{voc07}.  
For each dataset, in-domain and out-of-domain test sets are provided, with target objects listed below each context image.
  \textbf{D.} Top-1 accuracy of human participants and AI models on the lift-the-flap task under normal context conditions using the FRINE dataset. From left to right: SUP humans (black), SSL humans (gray), SeCo (red), self-supervised learning (SSL) baselines including ORL \cite{xie2021unsupervised}, SimSiam \cite{simsiam}, VICReg \cite{vicreg}, DINO \cite{dino}, SimCLR \cite{simclr}, and Context Encoder \cite{contextencoder} (orange), and a supervised learning baseline (green). A total of 517 SUP human trials, 548 SSL human trials, and 1,926 trials per model were collected. Error bars indicate standard errors computed across all trials for each group.
  \textbf{E.} Top-1 accuracy of human participants and SeCo under normal context conditions for Rules 1–3 of the FRINE dataset. Number of trials per rule is indicated in brackets: Rule 1—SUP humans (165), SSL humans (186); Rule 2—SUP humans (180), SSL humans (187); Rule 3—SUP humans (172), SSL humans (175). SeCo was evaluated on 1,926 trials for each rule. 
  \textbf{F.} Top-1 accuracy under normal context conditions for SUP humans, SSL humans, and SeCo across different context associations in the FRINE dataset. For each association type, the number of trials exceeds 640 for SUP humans, SSL humans, and SeCo. See \textbf{Methods} for definitions of the three context association types. Across \textbf{D}, \textbf{E}, and \textbf{F}, the chance level (25\%) is indicated by a horizontal black dashed line. Error bars indicate the standard error of the mean (SEM). * denotes performance significantly above chance based on Welch's two-tailed t-test ($p < 0.05$).
  }
  \end{minipage}
\end{figure}

\newpage

\newpage
\begin{figure}[!ht]
\begin{center}
\includegraphics[width=14.5cm]{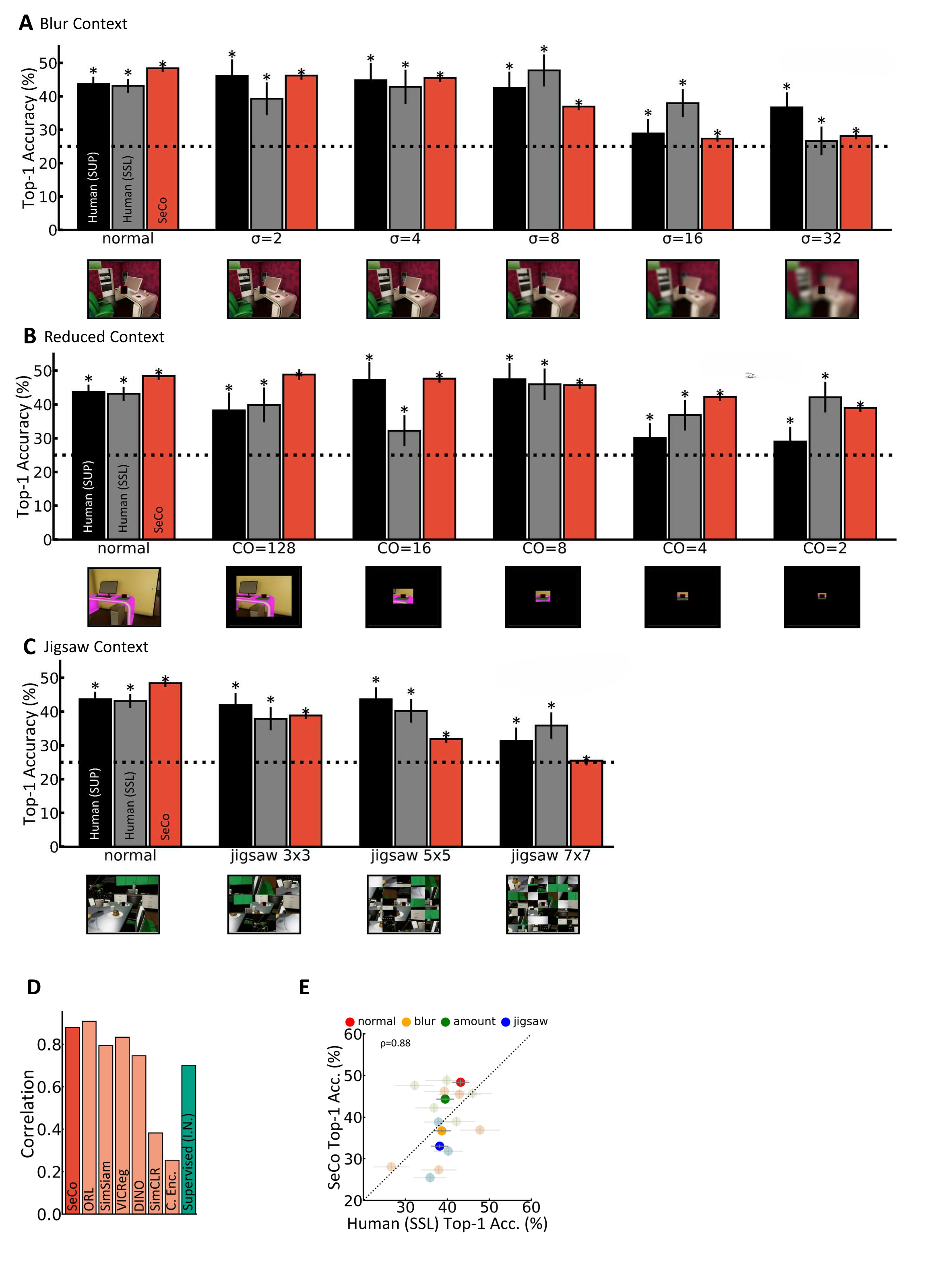}\vspace{-4mm}
\end{center}
\caption{
} 
\vspace{-5mm}
\label{fig:fig4cond}
\end{figure}

\begin{figure}
  \ContinuedFloat
  \caption{
\footnotesize 
\textbf{Contextual variations impact human and SeCo performance on the FRINE dataset in the lift-the-flap task.}  
\textbf{A.} Top-1 accuracy of humans and SeCo under the blur context condition. Context resolution was manipulated using zero-mean Gaussian blur with standard deviation $\sigma = 2, 4, 8, 16, 32$ pixels. Example images for each sub-condition are shown below the corresponding bar group. Format and design conventions follow \textbf{Fig.\ref{fig:fig2arch}E}.  
\textbf{B.} Top-1 accuracy of humans and SeCo under the reduced context area condition. Context amount was varied by adjusting the context-object ratio (CO) = 128, 16, 8, 4, 2. Format and conventions follow \textbf{A}.  
\textbf{C.} Top-1 accuracy of humans and SeCo under the jigsaw context condition. Spatial configuration of context was manipulated by randomly scrambling the image into 3$\times$3, 5$\times$5, or 7$\times$7 pieces, while the piece containing the target object remained in place. Format and conventions follow \textbf{A}.
\textbf{D.} Linear correlation between top-1 accuracy of SSL humans and AI models across all four context manipulations.  
\textbf{E.} Top-1 accuracy of SeCo versus SSL humans across four context manipulations: normal context (red), blur context (orange), reduced context area (green), and jigsaw context (blue). High-opacity dots indicate mean accuracy for each manipulation, while low-opacity dots of the same color show mean accuracy for sub-conditions within each manipulation. Cross-shaped error bars denote SEM across all trials. Pearson correlation coefficient: $\rho=0.88$. Black dotted line indicates the diagonal. 
Across \textbf{A}, and \textbf{B}, the chance level (25\%) is indicated by a horizontal black dashed line. Error bars indicate the standard error of the mean (SEM). * denotes performance significantly above chance based on Welch's two-tailed t-test ($p < 0.05$).
}
\end{figure}

\newpage
\begin{figure}[!ht]
\begin{center}
\includegraphics[width=15cm]{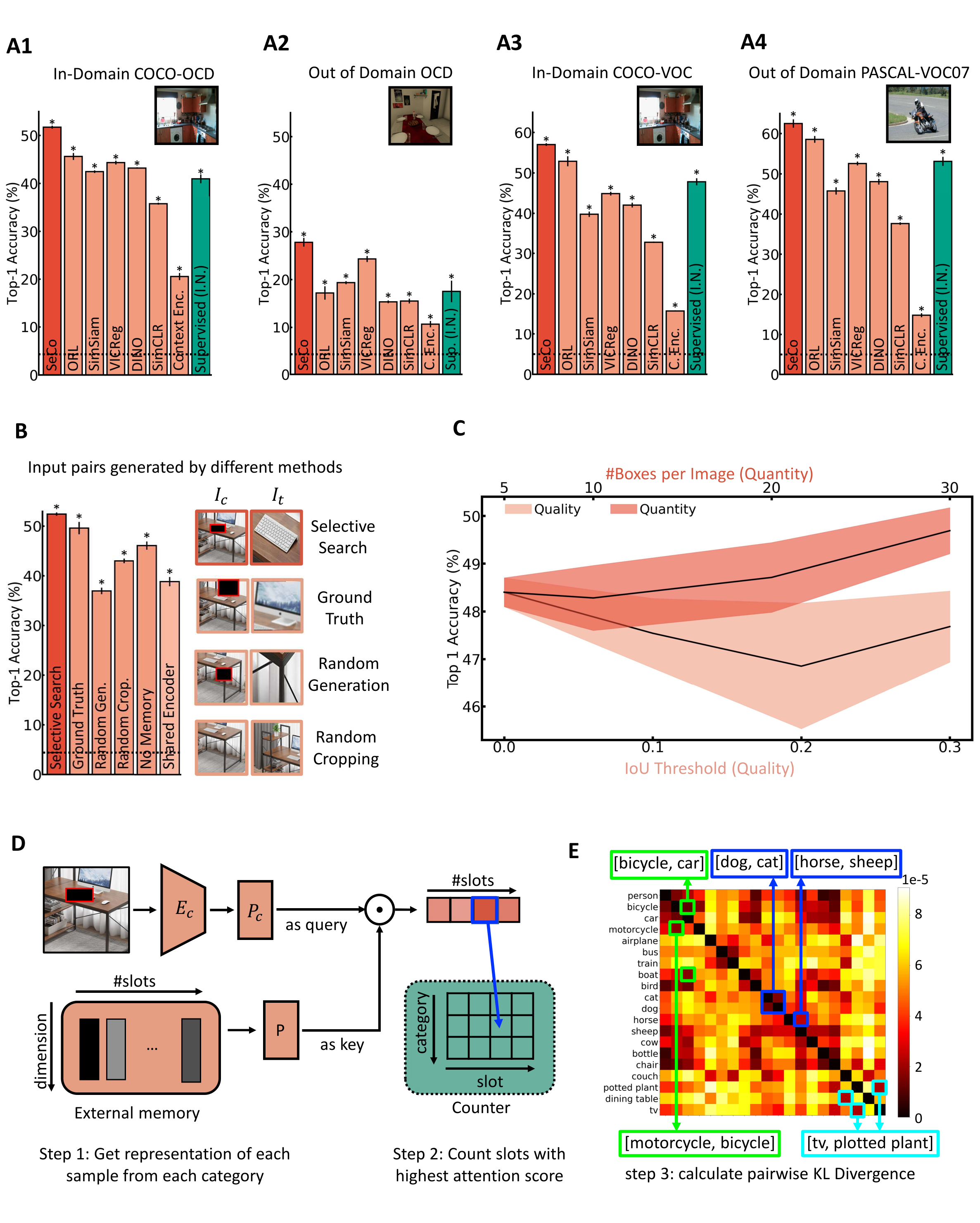}\vspace{-4mm}
\end{center}
\caption{
} 
\vspace{-5mm}
\label{fig:fig3analysis}
\end{figure}

\begin{figure}
  \ContinuedFloat
  \caption{
\footnotesize 
\textbf{SeCo outperforms all competitive baselines in the normal context in the lift-the-flap task, and model analysis reveals key architectural insights. 
}
\textbf{A.} Top-1 accuracy on the lift-the-flap task using the COCO-OCD and COCO-VOC datasets. Results are reported separately for in-domain (left) and out-of-domain (right) test sets. Representative examples from each test set are shown for illustration. Format and design conventions follow \textbf{Fig.~\ref{fig:fig2arch}E}.
\textbf{B.} Selective Search identifies effective context–object pairs for pre-training. To evaluate its contribution, we replaced SeCo’s context–object discovery module with: ground-truth bounding boxes as targets $I_t$ (\textit{Ground Truth}); randomly sampled crops as targets $I_t$ (\textit{Random Generation}); or two independent random crops, commonly used in self-supervised learning to form positive pairs $I_c$ and $I_t$ (\textit{Random Cropping}). Their performance is evaluated on the in-domain COCO-OCD dataset. Top-1 accuracy is reported alongside example context–object pairs generated by each variant, with frame colors corresponding to the bars in the plot.
Across \textbf{A}, and \textbf{B}, the chance level (6.7\%) is indicated by a horizontal black dashed line. Error bars indicate the standard error of the mean (SEM). * denotes performance significantly above chance based on Welch's two-tailed t-test ($p < 0.05$).
\textbf{C.} Proposal quantity has a slightly greater impact than proposal quality in SeCo. Top-1 accuracy on the in-domain COCO-OCD dataset is shown as a function of proposal quality (light orange; IoU threshold on bottom x-axis) and proposal quantity (dark orange; number of proposals on top x-axis). For the quality analysis, the number of proposals per image is fixed while varying the IoU threshold. For the quantity analysis, the IoU threshold is fixed at 0 while varying the number of proposals. Shaded regions denote standard error.
\textbf{D.} To examine what SeCo encodes in its external memory, we use test samples from the in-domain test set of COCO-VOC dataset and extract latent representations using the pre-trained context encoder $E_c$ (trapezoid). Attention scores over memory slots (squircles) are computed, and for each sample, the slot with the highest attention weight is identified (blue arrow).
\textbf{E.} Category-wise distributions of most-activated memory slots are aggregated across all samples, and pairwise Kullback–Leibler (KL) divergence is computed and visualized as a heatmap. Low divergence (dark cells) indicates that categories sharing similar contexts attend to similar memory slots. Colored boxes and arrows denote COCO-VOC supercategories: Vehicle (green), Animal (dark blue), and Indoor (cyan). For example, ``bicycle'' and ``car'' within the Vehicle supercategory activate overlapping memory regions, reflecting shared contextual associations.
}
\end{figure}

\newpage
\begin{figure}[!ht]
\begin{center}
\includegraphics[width=15cm]{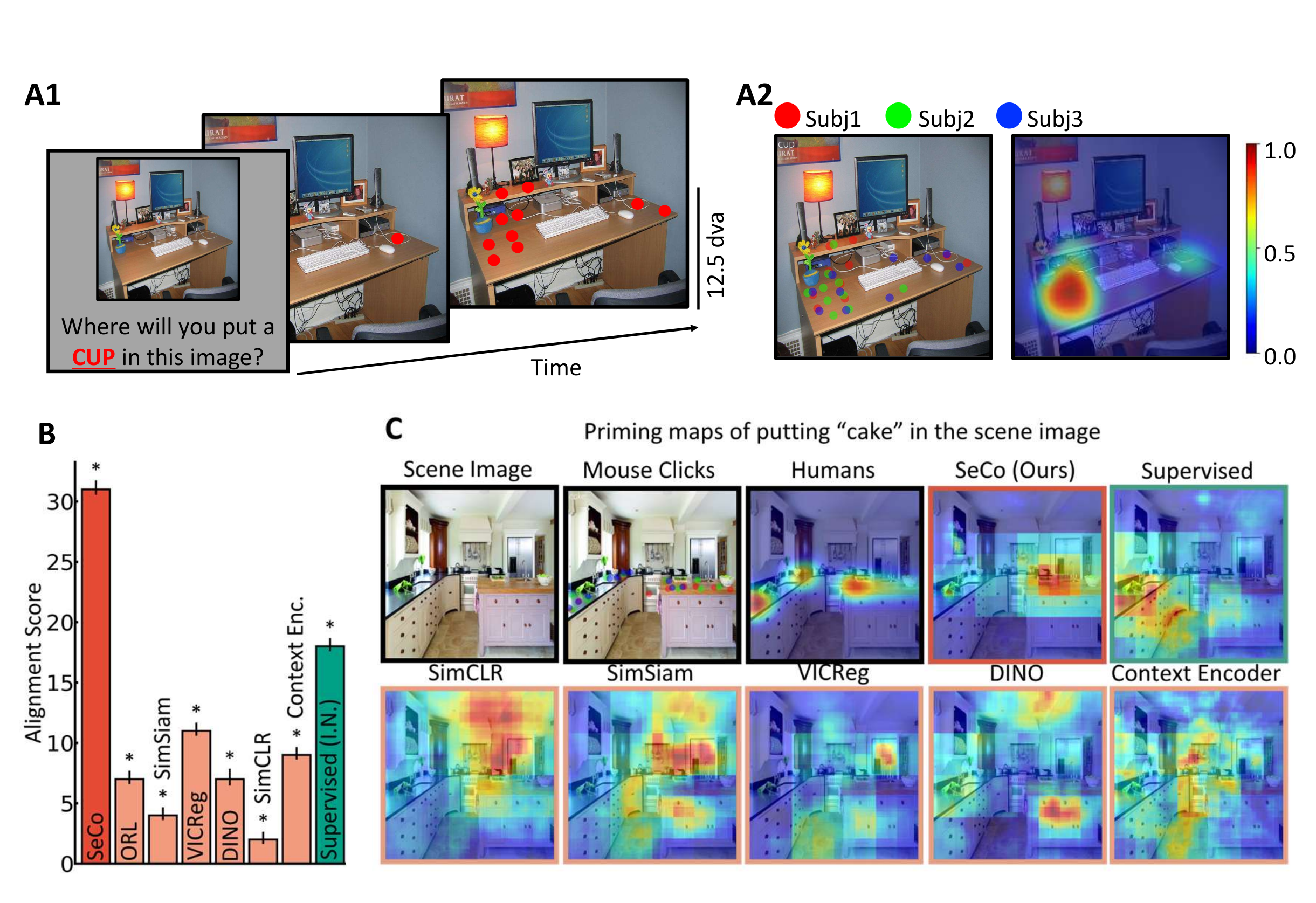}\vspace{-4mm}
\end{center}
\caption{\footnotesize \textbf{SeCo excels in the object priming task.}
\textbf{A.} \textbf{Schematic for human psychophysics experiments in the object priming task.}  
\textbf{A1.} Participants were first shown a natural image along with a target object. They were then instructed to place the object in appropriate locations by making 10 non-overlapping mouse clicks (indicated as red dots). There was no time limit for each trial. The scale bar on the right indicates the image size in degrees of visual angle.
 \textbf{A2.} Example priming map based on mouse clicks from humans. The left image shows the mouse click locations made by three human participants (colored dots) for \textit{cup} as the target object. The right panel shows the corresponding normalized probabilistic priming map generated from the consolidated clicks. A higher density of clicks corresponds to a higher probability in the priming map. See the color bar on the right for probability values.
\textbf{B.} 
Alignment scores between human priming maps and those predicted by SeCo (red), SSL baselines (orange), and a supervised baseline (green) are presented in the object-priming task. See \textbf{Methods} for the definition of alignment scores. A chance model yields a score of 0. Higher scores indicate better alighment with human clicks. Error bars show the standard error of the mean across five independent runs per model. * denotes performance significantly above chance based on Welch's two-tailed t-test ($p < 0.05$).
\textbf{C.} SeCo priming maps highlight contextually relevant regions of the image and closely approximate human choices in the object priming task. The leftmost column of the first row shows the input scene image where human and computational models are required to put the cake. The rest of the priming map from left to right, up to down are the consolidated click map from 3 human subjects with 10 non-overlapping mouse clicks (dots) each in different colors, ground truth priming maps generated from consolidated human mouse clicks, and priming maps predicted by our SeCo (framed in red) and predicted by supervised baseline (framed in green) and SSL baselines (framed in orange). 
See the color bar in \textbf{A2} for probability values. 
   } 
\vspace{-5mm}
\label{fig:fig7priming}
\end{figure}
\clearpage
\section*{Supplementary Figures}
\renewcommand{\thesection}{S\arabic{section}}
\renewcommand{\thefigure}{S\arabic{figure}}
\renewcommand{\thetable}{S\arabic{table}}
\setcounter{figure}{0}
\setcounter{section}{0}
\setcounter{table}{0}

\begin{figure}[!h]
\begin{center}
\includegraphics[width=13cm]{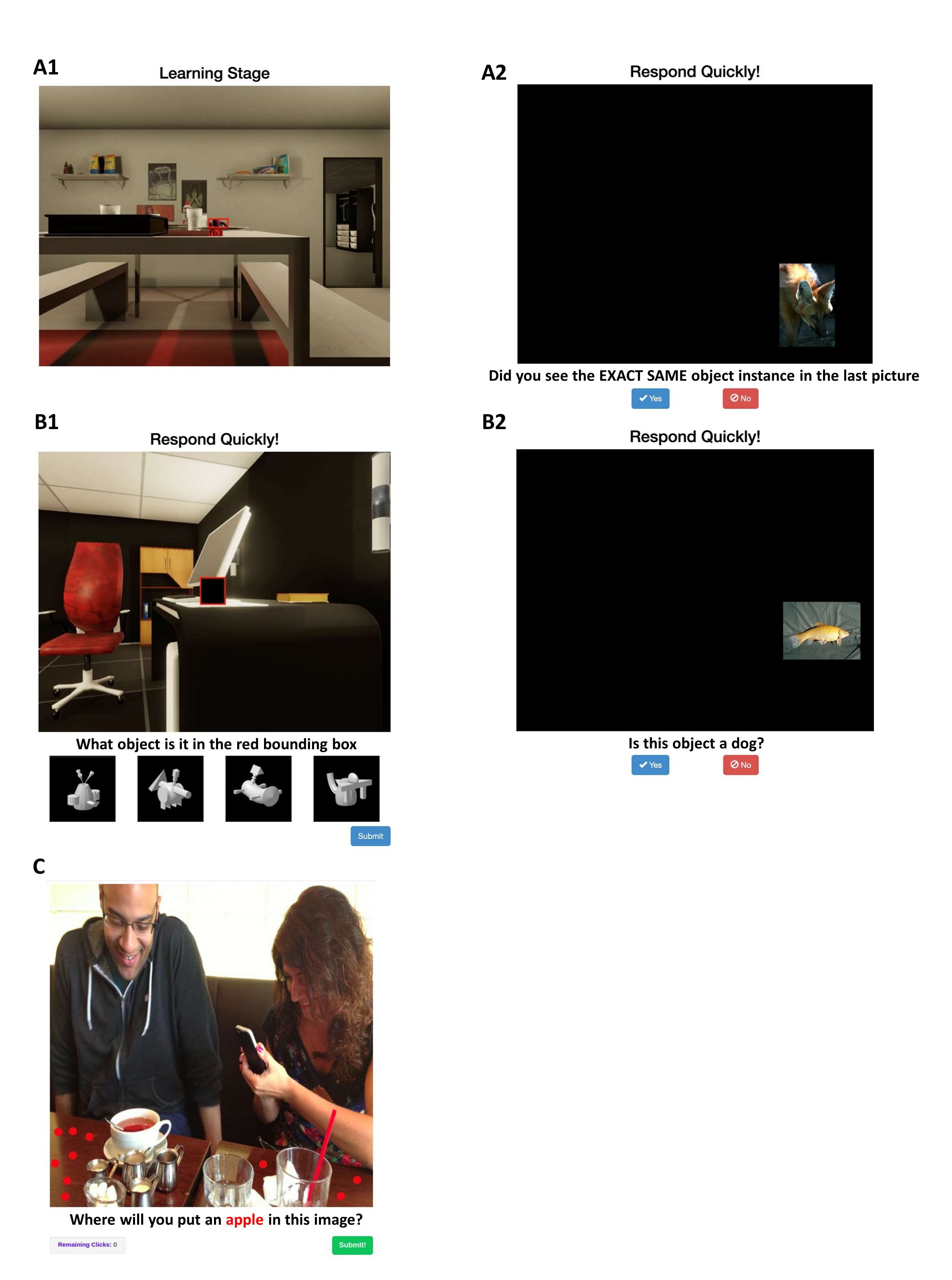}\vspace{-5mm}
\end{center}
\caption{ \footnotesize \textbf{Amazon Mechanical Turk (AMT) user interfaces of the psychophysics experiments.} 
\textbf{A1.} AMT user interface of the training stage from FRINE dataset. This example shows a screenshot when the human subject learns a fribble from family \textit{fb3} in the supervised learning condition.
\textbf{A2.} AMT user interface of memorization test in the training phase. In this example, an image of a fox was presented along with the question. The participant was required to decide whether they had seen this object in the current training video.
\textbf{B1.} AMT user interface of the lift-the-flap task in the test stage. A test video where the fribble was hidden under the black patch was presented along with the lift-the-flap question and four answer choices. The participant was expected to infer the identity of the hidden target and choose one of four fribble families as their answer.
\textbf{B2.} AMT user interface of the control trials in the test stage. This example shows an image of a fish and the participant was required to decide whether it is a dog.
\textbf{C.}  AMT user interface of object priming task. Participants are presented with various image-object pairs. For each pair, participants were required to make 10 non-repeated mouse clicks at relevant regions of the image. In this example, the participant was required to place an apple in the given scene. Red dots indicate the past click locations of the participant.
} 
\label{fig:figsscreenshots}
\vspace{-6mm}

\end{figure}
\newpage

\begin{figure}[!h]
\begin{center}
\includegraphics[width=12cm]{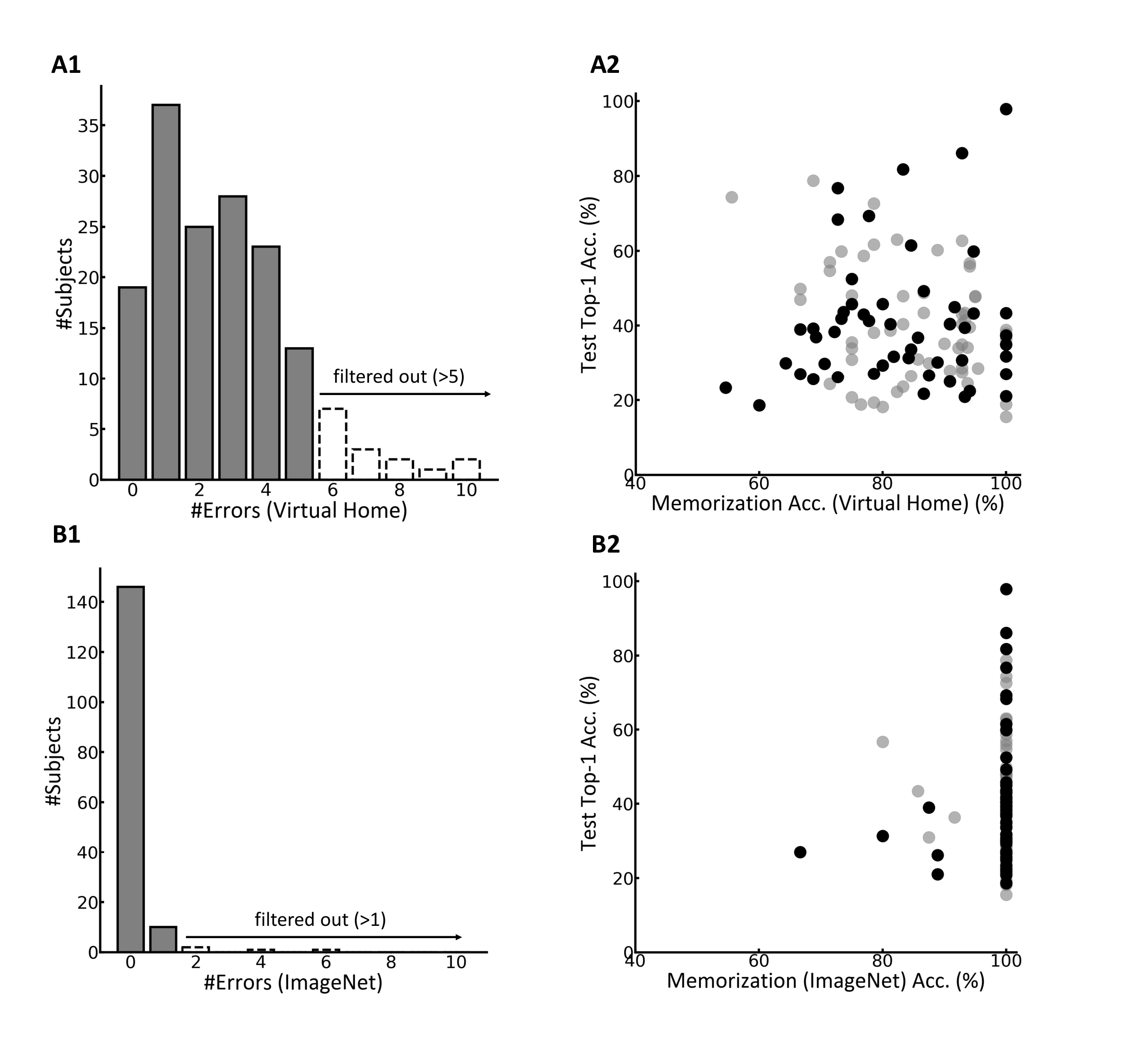}\vspace{-5mm}
\end{center}
\caption{\footnotesize \textbf{Quality of human subject data is high, with most participants passing control trials with high accuracy.}  
We report performance on the memorization test in control trials during the training phase of the human psychophysics experiment on the FRINE dataset.  
\textbf{A1.} Error distribution in the memorization tests where test objects were either positive samples from the current training video clip of FRINE (VirtualHome-Present) or negative samples from other FRINE training videos but absent from the current clip (VirtualHome-Absent). See \textbf{Methods} for details. Data from participants who made more than five errors (indicated by dashed white bars) were excluded.  
\textbf{A2.} Memorization test accuracy (VirtualHome-Present and VirtualHome-Absent) versus top-1 accuracy on the lift-the-flap task on the FRINE dataset. Each dot represents one participant trained via supervised learning (black) or self-supervised learning (gray).  
\textbf{B1.} Error distribution in the memorization test where test objects were always negative and they were sampled from ImageNet images whose class labels never appear in FRINE (ImageNet-Absent). See \textbf{Methods} for details. Data from participants who made more than one error (indicated by dashed white bars) were excluded.  
\textbf{B2.} Memorization test accuracy (ImageNet-Absent) versus top-1 accuracy on the lift-the-flap task using the FRINE dataset. Each dot represents one participant trained via supervised learning (black) or self-supervised learning (gray).  
}
\vspace{-6mm}
\label{fig:figs-memorization}
\end{figure}


\newpage
\begin{figure}[!h]
\begin{center}
\includegraphics[width=\linewidth]{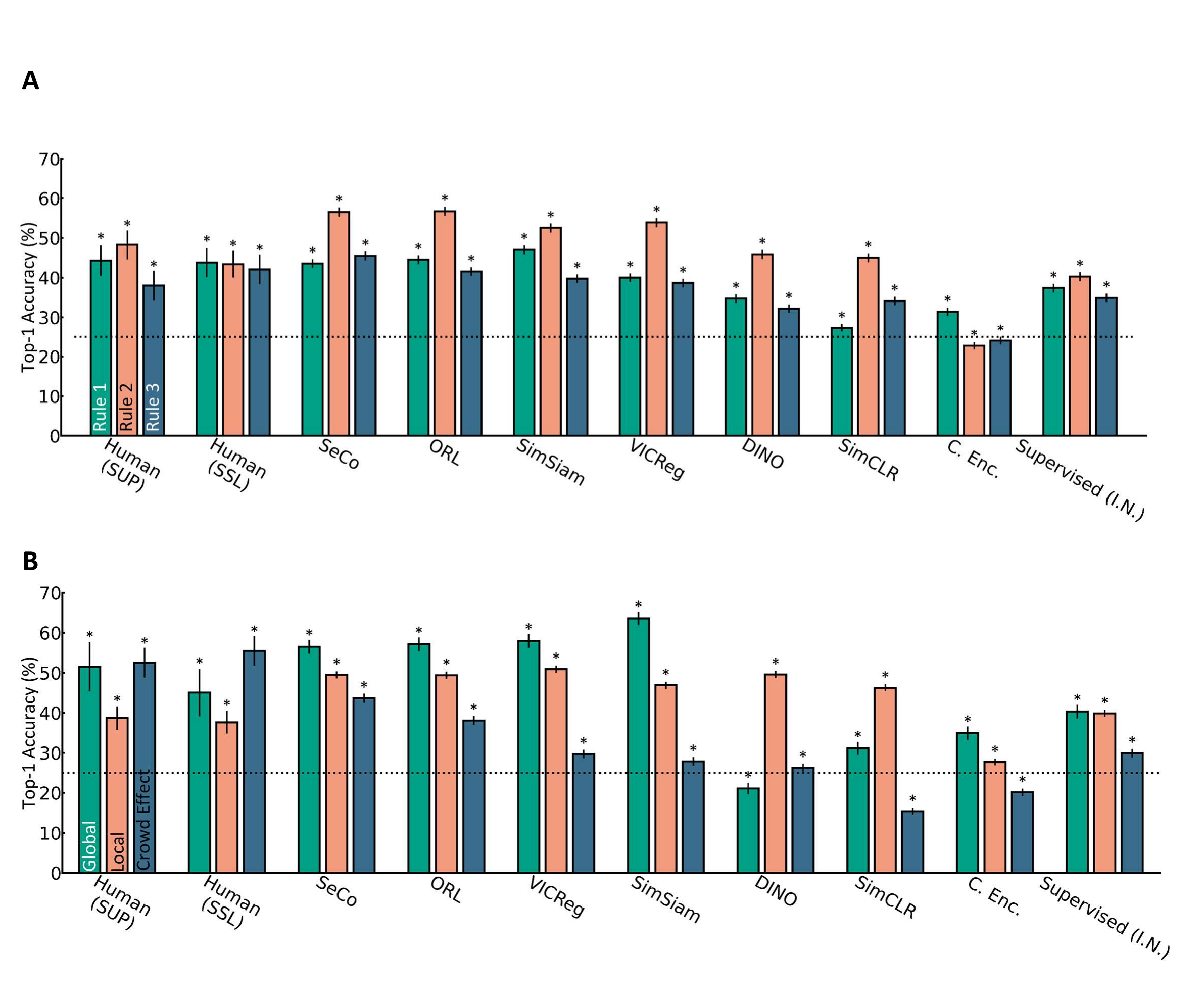}\vspace{-5mm}
\end{center}
\caption{\footnotesize
\textbf{The performance of humans and AI models on all three context rules and all three contextual associations on FRINE dataset.}
Top-1 accuracy of human participants and AI models in the lift-the-flap task under each \textbf{A. }contextual rule and \textbf{B. }type of contextual association on the FRINE dataset. From left to right, the models are ordered as follows: human participants trained via supervised learning, human participants trained via self-supervised learning, SeCo, ORL, VICReg, SimSiam, DINO, SimCLR, Context Encoder (C. Enc.), and the supervised model (Supervised (I.N.)). See \textbf{Methods} for details on all baselines. The chance-level performance is indicated by a horizontal black dotted line. The design convention follows that of \textbf{Fig.~\ref{fig:fig2arch}F}. 
}
\vspace{-6mm}
\label{fig:figs-glc}
\end{figure}

\newpage


\begin{figure}[!h]
\begin{center}
\includegraphics[width=15.5cm]{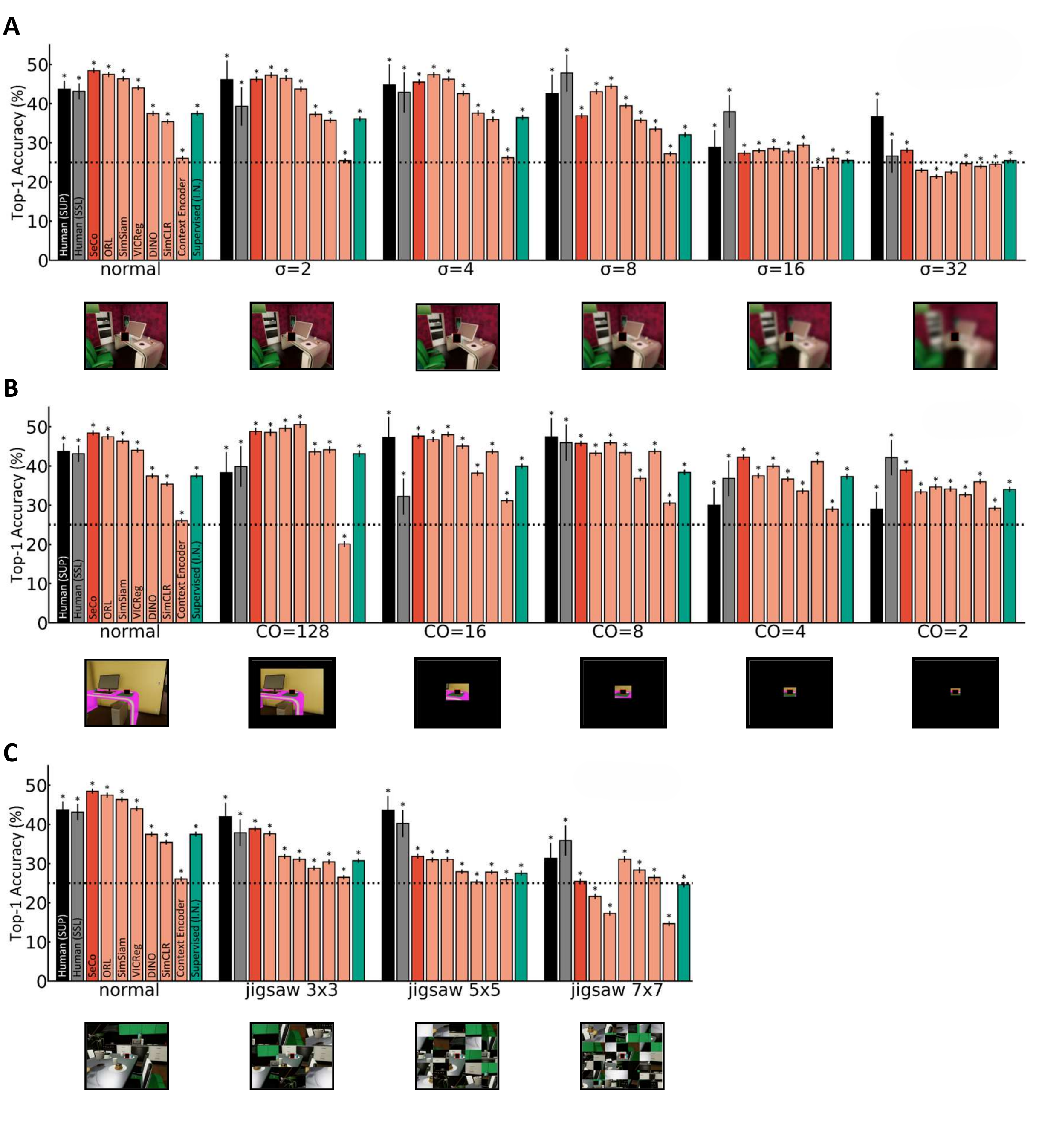}\vspace{-5mm}
\end{center}
\caption{\footnotesize 
\textbf{The effect of context on human and AI models in the lift-the-flap task on the FRINE dataset.}  
The design conventions in \textbf{A--C} follow those in \textbf{Fig.~\ref{fig:fig2arch}C--E}, respectively. See \textbf{Methods} for details on all baselines.
   } 
\vspace{-6mm}
\label{fig:figs-conditions}
\end{figure}

\newpage
\begin{figure}[!h]
\begin{center}
\includegraphics[width=10.5cm]{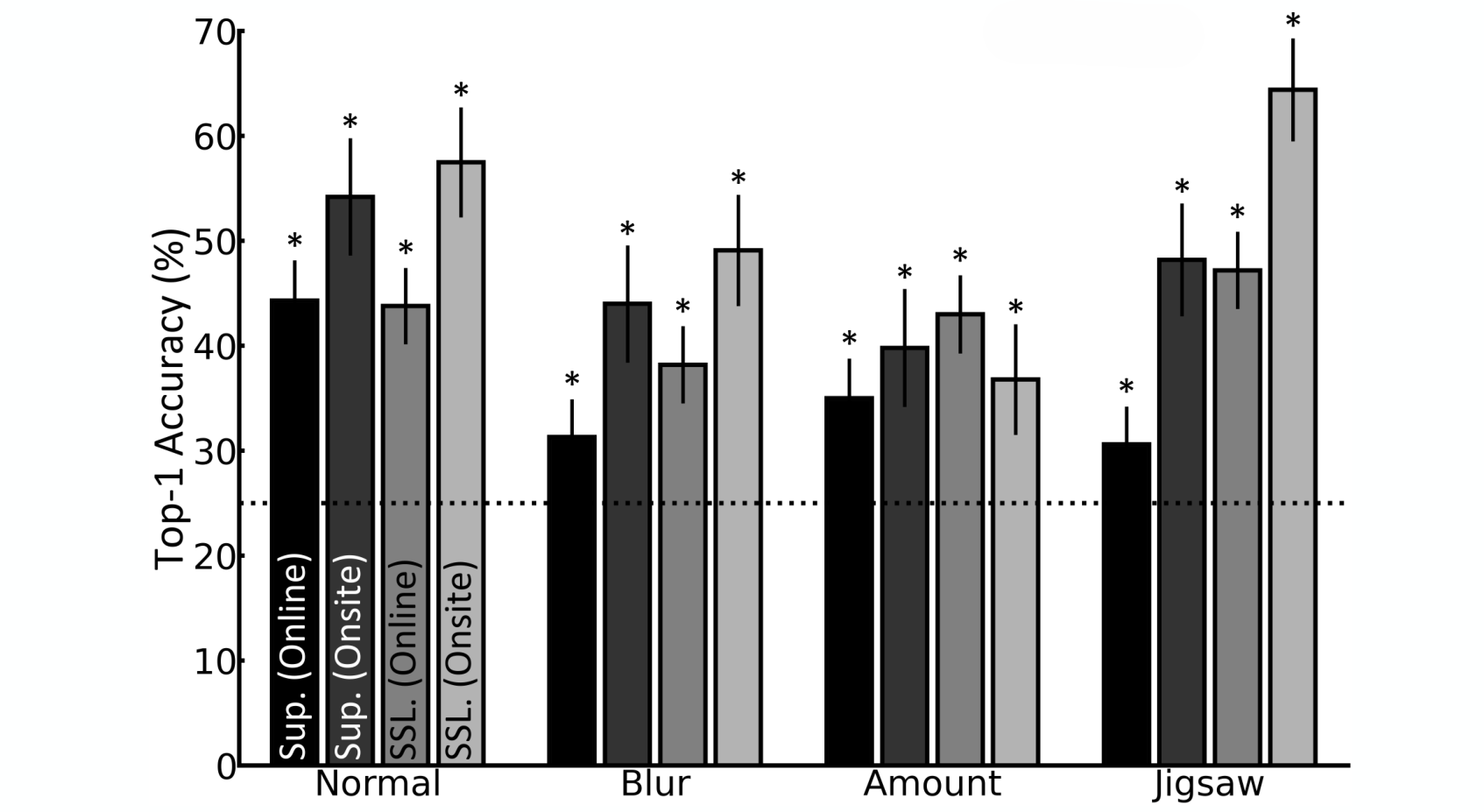}\vspace{-5mm}
\end{center}
\caption{\footnotesize  \textbf{Comparison of human performance in online and onsite experiments.} 
From left to right: top-1 accuracy of SUP humans in the online experiments (Sup. Online), SUP humans in the onsite experiment (Sup. Onsite), SSL humans in the online experiment (SSL. Online), SSL humans in the onsite experiment (SSL. Onsite). Results were grouped according to four context manipulations: normal, blur, amount, and jigsaw. The format and conventions follow those in \textbf{Fig.~\ref{fig:fig2arch}D}.
We recruited 20 onsite subjects (17 subjects left after the data quality control) to study the effect of prolonged exposure in LoR and the design of the curriculum on human performance of the lift-the-flap task on the FRINE dataset. See \textbf{Methods} for details on the onsite experiments. 
} 
\vspace{-6mm}
\label{fig:figsonsite}
\end{figure}

\newpage
\begin{figure}[!h]
\begin{center}
\includegraphics[width=15.5cm]{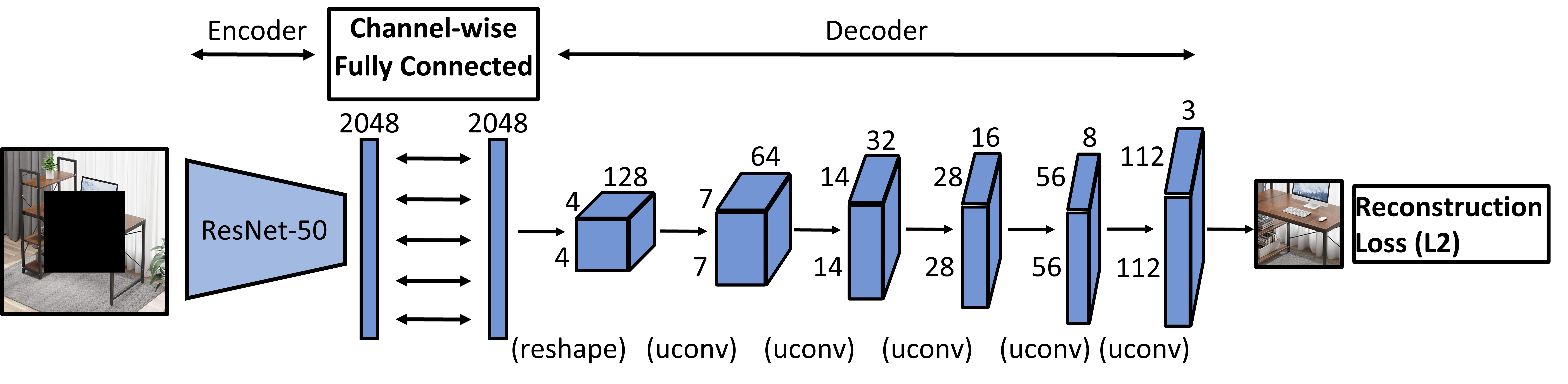}\vspace{-5mm}
\end{center}
\caption{\footnotesize  
\textbf{The architecture of Context Encoder \cite{contextencoder} with ResNet-50 \cite{resnet} as the backbone encoder.} To ensure fair comparisons with other SSL baselines, we replaced the AlexNet backbone used in the original Context Encoder with a ResNet-50 (trapezoid) architecture (see \textbf{Methods}). Following the original design, the encoder output is passed through a channel-wise fully connected layer (shown as double-headed arrows), followed by a five-layer decoder (represented by cuboids of varying sizes) that reconstructs the masked central region. We employed the same pixel-level L2 reconstruction loss as in the original configuration. 
}
\vspace{-6mm}
\label{fig:figs-contextenc}

\end{figure}



\newpage
\begin{figure}[!h]
\begin{center}
\includegraphics[width=10cm]{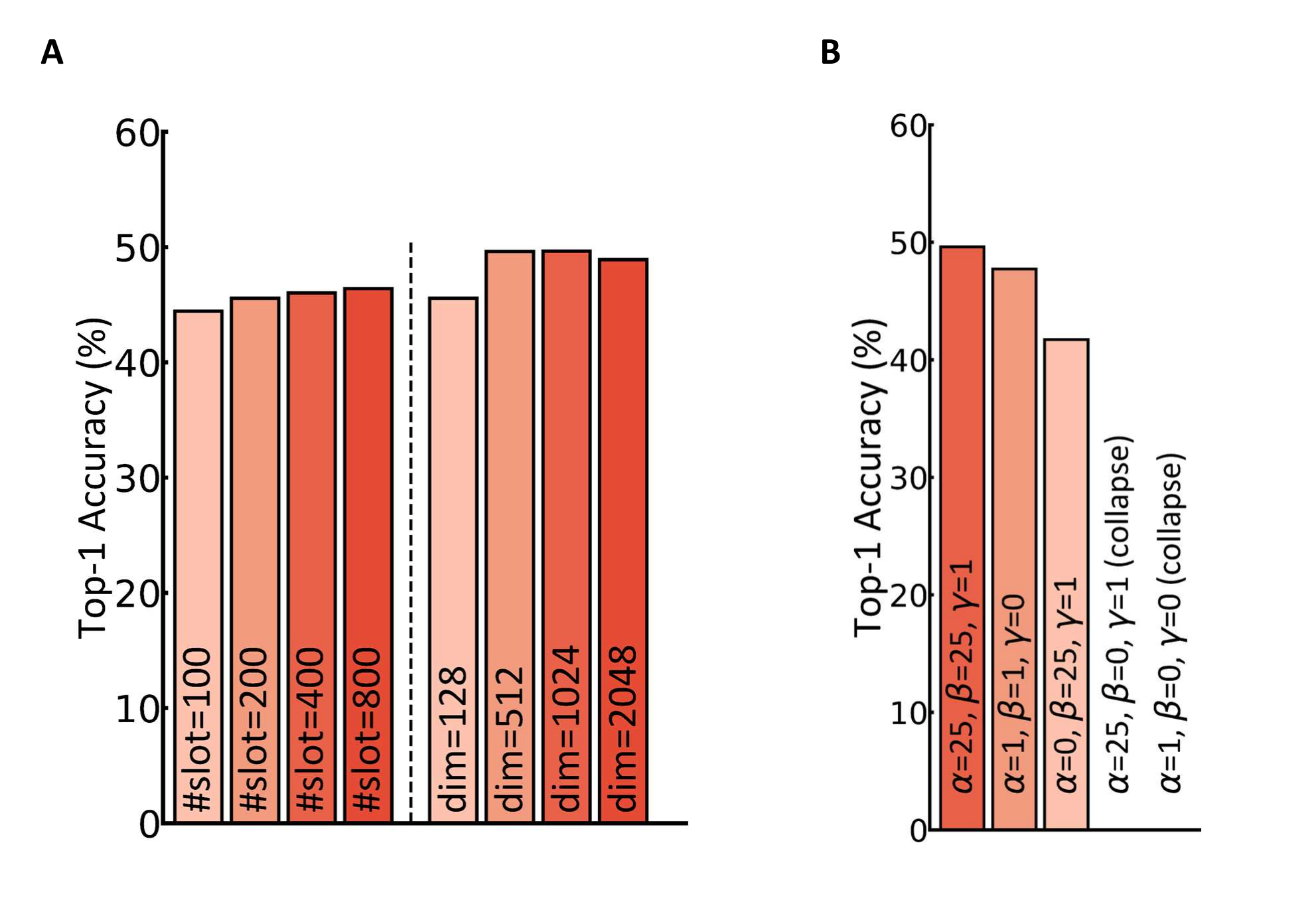}\vspace{-5mm}
\end{center}
\caption{\footnotesize  
\textbf{Impact of memory capacity (number of slots and memory vector dimension) and loss components on SeCo performance in the normal context on the COCO-OCD dataset in the lift-the-flap-task}. \textbf{A.} We analyze top-1 accuracy as a function of the number of memory slots ($K$) in the external memory module of SeCo and the vector dimension ($H$) for each memory slot. 
\textbf{B.} We also quantify the impact of individual loss components on SeCo’s reasoning performance (top-1 accuracy). The hyperparameters $\alpha$, $\beta$, and $\gamma$ represent the weights assigned to the Mean Squared Error (MSE) loss, variance loss, and covariance loss, respectively (Refer to \textbf{Eqn.~\ref{eqn:loss_eqn}}). Removing the variance loss component $\beta = 0$ results in model collapse. Hence, top-1 accuracy is not applicable in those cases.
}
\vspace{-6mm}
\label{fig:memory_slots}

\end{figure}


\newpage
\begin{figure}[!h]
\begin{center}
\includegraphics[width=11.5cm]{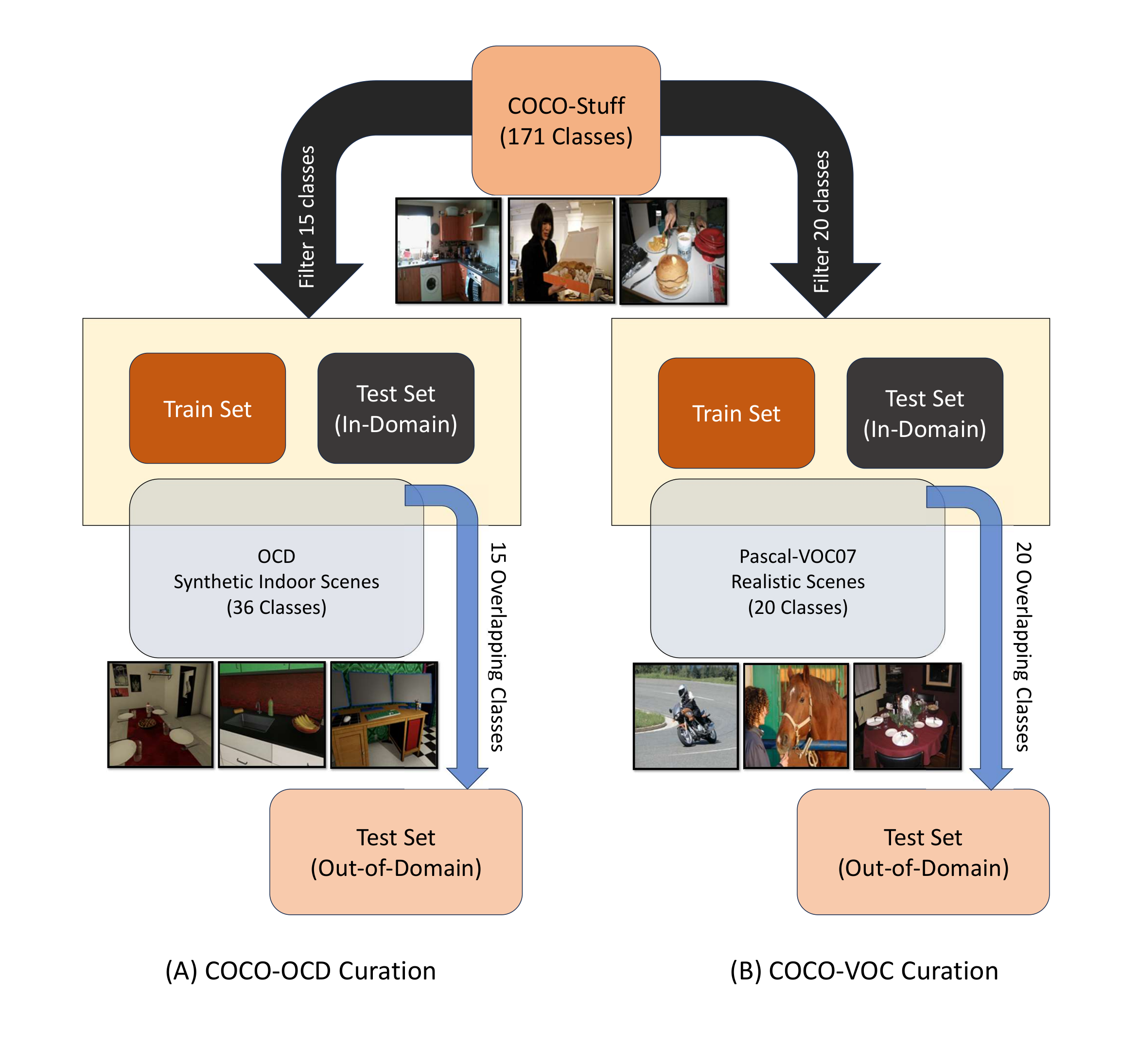}\vspace{-15mm}
\end{center}
\caption{\footnotesize  
\textbf{Schematic overview of the systematic curation of the COCO-OCD and COCO-VOC datasets}.
\textbf{(A)} COCO-OCD curation: Training and in-domain test sets were derived from COCO-Stuff \cite{cocostuff} by filtering for the 15 object classes shared with the Out-of-Context Dataset (OCD) \cite{whenpigsfly}. An out-of-domain test set was curated from OCD synthetic indoor scenes, restricted to the same 15 classes under normal context conditions.
\textbf{(B)} COCO-VOC curation: Following a procedure analogous to (A), training and in-domain test sets were derived from COCO-Stuff \cite{cocostuff} by filtering for the 20 object classes shared with the PASCAL-VOC07 dataset \cite{voc07}. The out-of-domain test set was curated from realistic PASCAL-VOC07 scenes containing the same 20 target classes.
}
\vspace{-6mm}
\label{fig:dataset}

\end{figure}

\end{document}